\documentclass[acmlarge]{acmart}

\usepackage{booktabs} 
\usepackage{threeparttable}
\usepackage{multirow}
\usepackage[ruled, linesnumbered]{algorithm2e} 
\usepackage[normalem]{ulem}

\SetAlFnt{\small}
\SetAlCapFnt{\small}
\SetAlCapNameFnt{\small}
\SetAlCapHSkip{0pt}
\IncMargin{-\parindent}

\acmJournal{TALLIP}
\acmVolume{17}
\acmNumber{2}
\acmArticle{9}
\acmYear{2017}
\acmMonth{11}
\copyrightyear{2017}
\acmArticleSeq{9}

\setcopyright{acmlicensed}

\acmDOI{https://doi.org/10.1145/3138815}

\received{February 2017}
\received[Major Revision]{May 2017}
\received[Minor Revision]{August 2017}
\received[accepted]{June 2009}

\begin{document}
	\title[A Generalized Constraint Approach to Bilingual Dictionary Induction]{A Generalized Constraint Approach to Bilingual Dictionary Induction for Low-Resource Language Families} 
	\author{Arbi Haza Nasution}
	\orcid{0000-0001-6283-3217}
	\affiliation{%
		\institution{Kyoto University}
		\department{Social Informatics}
		\city{Kyoto}
		\country{Japan}
	}
	\affiliation{%
		\institution{Universitas Islam Riau}
		\department{Information Technology}
		\city{Pekanbaru}
		\country{Indonesia}
	}
	\author{Yohei Murakami}
	\affiliation{%
		\institution{Kyoto University}
		\department{Unit of Design}
		\city{Kyoto}
		\country{Japan}
	}
	\author{Toru Ishida}
	\affiliation{%
		\institution{Kyoto University}
		\department{Social Informatics}
		\city{Kyoto}
		\country{Japan}
	}
	\begin{abstract}
		The lack or absence of parallel and comparable corpora makes bilingual lexicon extraction a difficult task for low-resource languages. The pivot language and cognate recognition approaches have been proven useful for inducing bilingual lexicons for such languages. We propose constraint-based bilingual lexicon induction for closely-related languages by extending constraints from the recent pivot-based induction technique and further enabling multiple symmetry assumption cycle to reach many more cognates in the transgraph. We further identify cognate synonyms to obtain many-to-many translation pairs. This paper utilizes four datasets: one Austronesian low-resource language and three Indo-European high-resource languages. We use three constraint-based methods from our previous work, the Inverse Consultation method and translation pairs generated from Cartesian product of input dictionaries as baselines. We evaluate our result using the metrics of precision, recall and F-score. Our customizable approach allows the user to conduct cross validation to predict the optimal hyperparameters (cognate threshold and cognate synonym threshold) with various combination of heuristics and number of symmetry assumption cycles to gain the highest F-score. Our proposed methods have statistically significant improvement of precision and F-score compared to our previous constraint-based methods. The results show that our method demonstrates the potential to complement other bilingual dictionary creation methods like word alignment models using parallel corpora for high-resource languages while well handling low-resource languages.		
	\end{abstract}

	%
	%

	%
	%
	
	
	
	\thanks{This paper is significantly extended from our previous work: \cite{NASUTION16.1238}\\
		A. H. Nasution is with Department of Social Informatics, Kyoto University, Kyoto, Japan, email: arbi@ai.soc.i.kyoto-u.ac.jp and Department of Information Technology, Universitas Islam Riau, Pekanbaru, Indonesia, email: arbi@eng.uir.ac.id.\\
		Y. Murakami is with Unit of Design, Kyoto University, Kyoto, Japan, email: yohei@i.kyoto-u.ac.jp. \\
		T. Ishida is with Department of Social Informatics, Kyoto University, Kyoto, Japan, email: ishida@i.kyoto-u.ac.jp.}
	
	\maketitle
	\renewcommand{\shortauthors}{A. H. Nasution et al.}
	\section{Introduction}
	
	Machine readable bilingual dictionaries are very useful for information retrieval and natural language processing research, but are usually unavailable for low-resource languages. Previous work on high-resource languages showed the effectiveness of parallel corpora \cite{Fung-98}\cite{brown1990statistical} and comparable corpora \cite{rapp1995identifying}\cite{fung1995compiling} in inducing bilingual lexicons. Bilingual lexicon extraction is highly problematic for low-resource languages due to the paucity or outright omission of parallel and comparable corpora. The approaches of pivot language \cite{tanaka-94} and cognate recognition \cite{Mann-01} have been proven useful in inducing bilingual lexicons for low-resource languages. Closely-related languages share cognates that share most of the semantic or meaning of the lexicons \cite{lehmann2013historical}. Some linguistics studies \cite{van2005easy}\cite{gooskens2006linguistic} show that the percentage of shared cognates, either related directly or via a synonym, constitutes a highly accurate linguistic distance measure based on mutual intelligibility, i.e. the ability of speakers of one language to understand the other language. The higher the percentage of shared cognates between the languages, the lower the linguistic distance, the higher is the level of mutual intelligibility.\par
	We recently introduced the promising approach of treating pivot-based bilingual lexicon induction for low-resource languages as an optimization problem \cite{NASUTION16.1238} with cognate pair coexistence probability as a sole heuristic in the symmetry constraint. In this paper, we propose generalized constraint-based bilingual lexicon induction for closely-related languages by setting two steps to obtaining translation pair results. First, we identify one-to-one cognates by incorporating more constraints and heuristics to improve the quality of the translation result. We then identify the cognates' synonyms to obtain many-to-many translation pairs. In each step, we can obtain more cognate and cognate synonym pair candidates by iterating the n-cycle symmetry assumption until all possible translation pair candidates have been reached. We address the following research goals:
	\begin{itemize}
		\item{\textit{Creating many-to-many translation pairs between closely-related languages}: Recognize cognates and cognate synonyms from direct and indirect connectivities via pivot word(s) by iterating the symmetry assumption cycle to improve the quality and quantity of the translation pair results.}
		\item{\textit{Evaluating the generalized method performance}: We apply the Inverse Consultation method \cite{tanaka-94} and naive translation pairs generation from the Cartesian product of input dictionaries to all of our datasets and compare the results with those of our generalized methods using precision, recall and F-score. We also conduct experiments with our previous constraint-based methods \cite{NASUTION16.1238} with the same datasets and further conduct student's paired t-tests to show that our proposed methods have statistically significant improvement of precision and F-score. We conduct cross validation to predict the optimal hyperparameters (cognate threshold and cognate synonym threshold) to gain the highest F-score.}		
	\end{itemize}
	The rest of this paper is organized as follows: In Section 2, we will briefly discuss related research on bilingual dictionary induction. Section 3 discusses closely-related languages and existing methods in comparative linguistics. Section 4 details our strategy of recognizing cognate and cognate synonyms, core component for our proposal, which is described in Section 5. Section 6 introduces our experiment and the results. Finally, Section 7 concludes this paper.
	\section{Bilingual Dictionary Induction}
	An intermediate/pivot language approach has been applied in machine translation \cite{tanaka2009context} and service computing \cite{ishida2011language} researches. The first work on bilingual lexicon induction to create bilingual dictionary between language A and language C via pivot language B is Inverse Consultation (IC) \cite{tanaka-94} by utilizing the structure of input dictionaries to measure the closeness of word meanings and then use the results to prune erroneous translation pair candidates. The IC approach identifies equivalent candidates of language A words in language C by consulting dictionary A-B and dictionary B-C. These equivalent candidates will be looked up and compared in the inverse dictionary C-A. To analyze the method used to filter wrong translation pair candidates induced via the pivot-based approach, \cite{saralegi2011analyzing} explored distributional similarity measure (DS) in addition to IC. The analysis showed that IC depends on significant lexical variants in the dictionaries for each meaning in the pivot language, while DS depends on distributions or contexts across two corpora of the different languages. Their analysis also showed that the combination of IC and DS outperformed each used individually.\par
	The pivot-based approach is very suitable for low-resource languages, especially when dictionaries are the only language resource required. Unfortunately, for some low-resource languages, it is often difficult to find machine-readable inverse dictionaries and corpora to filter the wrong translation pair candidates. Thus, we consider that the combination of IC and DS methods does not suit low-resource languages. To overcome this limitation, our team \cite{wushouer-15} proposed to treat pivot-based bilingual lexicon induction as an optimization problem. The assumption was that lexicons of closely-related languages offer one-to-one mapping and share a significant number of cognates (words with similar spelling/form and meaning originating from the same root language). With this assumption, they developed a constraint optimization model to induce an Uyghur-Kazakh bilingual dictionary using Chinese language as the pivot, which means that Chinese words were used as intermediates to connect Uyghur words in an Uyghur-Chinese dictionary with Kazakh words in a Kazakh-Chinese dictionary. They used a graph whose vertices represent words and edges indicate shared meanings; they called this a transgraph following \cite{Soderland-09}. The steps in their approach are as follows: (1) use two bilingual dictionaries as input, (2) represent them as transgraphs where $w_1^A$ and $w_2^A$ are non-pivot words in language A, $w_1^B$ and $w_2^B$ are pivot words in language B, and $w_1^C$, $w_2^C$ and $w_3^C$ are non-pivot words in language C, (3) add some new edges represented by dashed edges based on the one-to-one assumption, (4) formalize the problem into conjunctive normal form (CNF) and use the Weighted Partial MaxSAT (WPMaxSAT) solver \cite{ansotegui2009solving} to return the optimized translation results, and (5) output the induced bilingual dictionary as the result. These steps are shown in Fig. 1.
	\begin{figure}[!h]
		\begin{center}
			\includegraphics[scale=0.6]{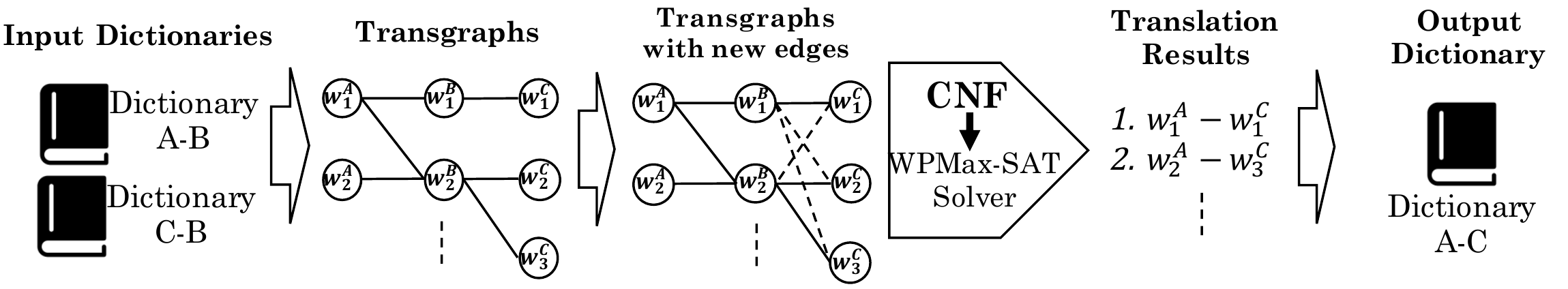} 
			\caption{One-to-one constraint approach to pivot-based bilingual dictionary induction.}
			\label{fig.1}
		\end{center}
	\end{figure}	
	The one-to-one approach depends only on semantic equivalence, one of the closely-related language characteristics that permit the recognition of cognates between languages assuming that lexicons of closely-related languages offer the one-to-one relation. If language A and C are closely related, for any word in A there exists a unique word in C such that they have exactly the same meaning, and thus are symmetrically connected via pivot word(s). Such a pair is called a one-to-one pair. They realized that this assumption may be too strong for the general case, but they believed that it was reasonable for closely-related languages like Turkic languages. They believe that their method works best for languages with high-similarity. They tried to improve the precision by utilizing multiple input dictionaries \cite{Wushouer2014} while still applying the same one-to-one assumption. However, this assumption is too strong to be used for the induction of as many translation pairs as possible to offset resource paucity because the few such pairs are yielded.
	
	\section{Closely Related Languages}
	Historical linguistics is the scientific study of language change over time in term of sound, analogical, lexical, morphological, syntactic, and semantic information \cite{campbell2013historical}. Comparative linguistics is a branch of historical linguistics that is concerned with language comparison to determine historical relatedness and to construct language families \cite{lehmann2013historical}. Many methods, techniques, and procedures have been utilized in investigating the potential distant genetic relationship of languages, including lexical comparison, sound correspondences, grammatical evidence, borrowing, semantic constraints, chance similarities, sound-meaning isomorphism, etc \cite{campbell2008language}. The genetic relationship of languages is used to classify languages into language families. Closely-related languages are those that came from the same origin or proto-language, and belong to the same language family.\par
	Automated Similarity Judgment Program (ASJP) was proposed by \cite{holman2011automated} with the main goal of developing a database of Swadesh lists \cite{swadesh1955towards} for all of the world's languages from which lexical similarity or lexical distance matrix between languages can be obtained by comparing the word lists. We utilize ASJP to select our low-resource target languages for our first case study in this paper. Indonesia has 707 low-resource ethnic languages \cite{Lewis-15} which are suitable as target languages in our study. There are three factors we consider in selecting the target languages: language similarity, input bilingual dictionary size, and number of speakers. In order to ensure that the induced bilingual dictionaries will be useful for many users, we listed the top 10 Indonesian ethnic languages ranked by the number of speakers. We then generated the language similarity matrix by utilizing ASJP as shown in Table 1. From this list, the biggest size machine readable bilingual dictionaries are Minangkabau-Indonesian and Malay-Indonesian. After considering all those factors, we selected Malay, Minangkabau and Indonesian as our target languages for the low-resource languages case study. \par
	\begin{table}
		\tiny
		\caption{Similarity Matrix of Top 10 Indonesian Ethnic Languages Ranked by Number of Speakers}
		\label{tab:1}
		\begin{center}
			\begin{tabular}{lccccccccc}
				\toprule
				Language&Indonesian&Malang&Yogyakarta&Old Javanese&Sundanese&Malay&Palembang Malay&Madurese&Minangkabau\\
				\hline
				Malang&23.46\%& & & & & & & &\\
				Yogyakarta&27.29\%&87.36\%&&&&&&&\\
				Old Javanese&24.09\%&47.50\%&52.18\%&&&&&&\\
				Sundanese&39.43\%&18.55\%&22.43\%&21.82\%&&&&&\\
				Malay&\textbf{85.10\%}&20.53\%&24.35\%&21.36\%&41.12\%&&&&\\
				Palembang Malay&68.24\%&33.97\%&37.97\%&31.85\%&38.90\%&73.23\%&&&\\
				Madurese&34.45\%&17.63\%&14.15\%&15.18\%&19.86\%&34.16\%&34.32\%&&\\
				Minangkabau&\textbf{61.59\%}&26.59\%&29.63\%&25.01\%&30.81\%&\textbf{61.66\%}&63.60\%&34.32\%&\\
				Buginese&31.21\%&12.76\%&16.85\%&18.33\%&24.80\%&32.04\%&31.00\%&17.94\%&32.00\%\\
				\bottomrule
			\end{tabular}
		\end{center}
	\end{table}
	Several machine translation studies focused on closely-related languages \cite{scannell2006machine,nakov2012combining,tiedemann2009character}. In this research, the linguistic characteristics of the closely-related languages play a vital role in improving quality of our method.
		
	\section{Cognate and Cognate Synonym Recognition}
	\begin{figure}[!h]
		\begin{center}
			\includegraphics[scale=0.8]{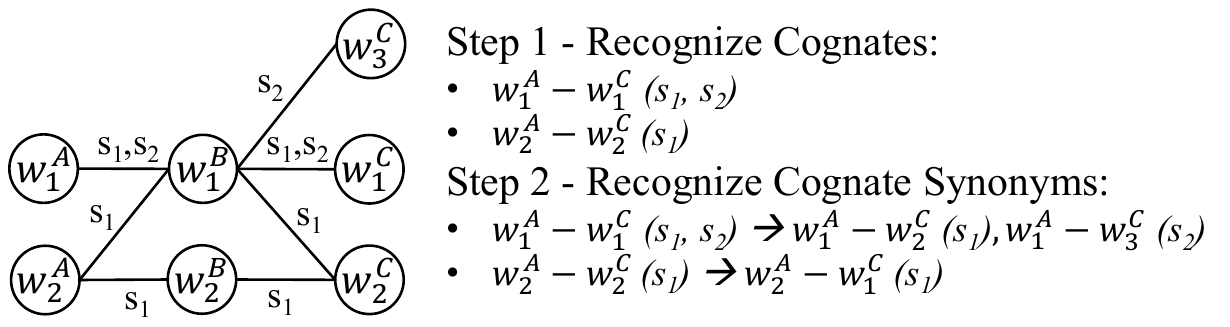} 
			\caption{Strategy to recognize cognates and cognate synonyms.}
			\label{fig.2}
		\end{center}		
	\end{figure}
	By utilizing linguistic information, we establish a strategy to obtain many-to-many translation pairs from a transgraph. The first step is to recognize one-to-one cognates in the transgraph which share all their senses. Once a list of cognates is obtained, the next step is to recognize cognate synonyms in the transgraph; those that share part/all senses with the cognate and so are mutually connected to some/all pivot words. Those two steps are easy tasks when the input dictionaries have sense/meaning information as shown in Fig. 2 where a cognate pair $(w_1^A, w_1^C)$ share two senses, i.e., $s_1$ and $s_2$ through pivot word $w_1^B$ and a cognate pair $(w_2^A, w_2^C)$ only share $s_1$ through pivot word $w_1^B$ and $w_2^B$. Since for low-resource languages, a machine-readable bilingual dictionary with sense information is scarce, we regard connected words share at least one sense/meaning. Thus, we assume that non-pivot words which are symmetrically connected via pivot word(s) potentially share all their senses and so being a cognate. \par
	Cognates are words with similar spelling/form and meaning that have a common etymological origin. For instance, the words \textit{night} (English), \textit{nuit} (French), \textit{noche} (Spanish), \textit{nacht} (German) and \textit{nacht} (Dutch) have the same meaning which is "night" and derived from the Proto-Indo-European \textit{*n\'{o}$k^w$ts} with the same meaning of "night".
	Since most linguists believe that lexical comparison alone is not a good way to recognize cognates \cite{campbell2013historical}, we want to utilize a more general and basic characteristic of closely-related languages, which is: a cognate pair mostly maintain the semantic or meaning of the lexicons. Even though there is a possibility of a change in one of the meanings of a word in a language, within the families where the languages are known to be closely-related, the possibility of a change is smaller. Since our approach targets the closely-related languages, it is safe to make the following assumption based on the semantic characteristic of closely-related languages: \textit{Given a pair of words, $w_i^A$ of language A and $w_k^C$ of language C, if they are cognates, they share all of their senses/meanings and are symmetrically connected through pivot word(s) from language B.} We call this the symmetry assumption. Unfortunately, in some cases, symmetry assumption is inadequate to eliminate wrong cognate from the cognate pair candidates when a pivot-word has multiple indegree/outdegree. To correctly find cognates, not only the meaning (which is predicted by shared edges), but also the form need to be considered. We add form-similarity/lexical distance rate as a new heuristic in finding cognates following \cite{melamed1995induce} using the Longest Common Subsequence Ratio (LCSR). \par
	\begin{figure}[!h]
		\begin{center}
			\includegraphics[scale=0.9]{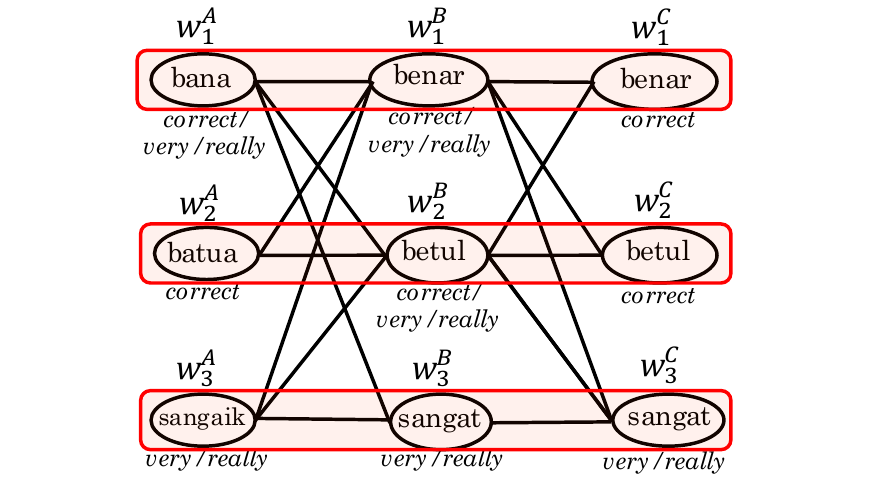} 
			\caption{Cognate and cognate synonym example.}
			\label{fig.3}
		\end{center}
	\end{figure}
	Some linguistic studies show that the meaning of a word can be deduced via cognate synonym \cite{van2005easy,gooskens2006linguistic}. For instance, in Fig. 3, $w_1^A$, $w_2^A$ and $w_3^A$ are words in Minangkabau language (min), $w_1^B$, $w_2^B$ and $w_3^B$ are words in Indonesian language (ind) and $w_{1}^C$, $w_{2}^C$ and $w_{3}^C$ are words in Malay language (zlm). When we connect words in non-pivot language A and C via pivot words B based on shared meaning between the words, we can get translation results from language A to C.  In this example, we have information about senses/meanings for all words in input dictionaries and there are three cognates which are $(w_1^A, w_1^B, w_1^C)$, $(w_2^A, w_2^B, w_2^C)$, and $(w_3^A, w_3^B, w_3^C)$ as indicated within the same box in Fig. 3. A cognate $w_1^A-w_1^C$ and non-cognates $w_1^A-w_2^C$ and $w_1^A-w_3^C$ are correct translations since $w_1^C$, $w_2^C$ and $w_3^C$ are synonymous.\par
	Nevertheless, it remains a challenge to find the cognate synonyms when the input dictionaries do not have information about senses/meanings. As shown in Fig. 4, to recognize cognate synonyms, firstly, we need to recognize synonyms of $w_2^C$ based on ratio of shared connectivity with the pivot word(s), since we assume that synonymous words are connected to common pivot word(s). Then, $w_1^A$ will be paired with the recognized synonyms of $w_2^C$ to obtain cognate synonym pairs. The higher the ratio of shared connectivity between a synonym of $w_2^C$ with the pivot words ($w_1^B, w_2^B, w_3^B$), the higher the probability of the synonym being a translation pair with $w_1^A$.\par
	\begin{figure}[!h]
		\begin{center}
			\includegraphics[scale=0.45]{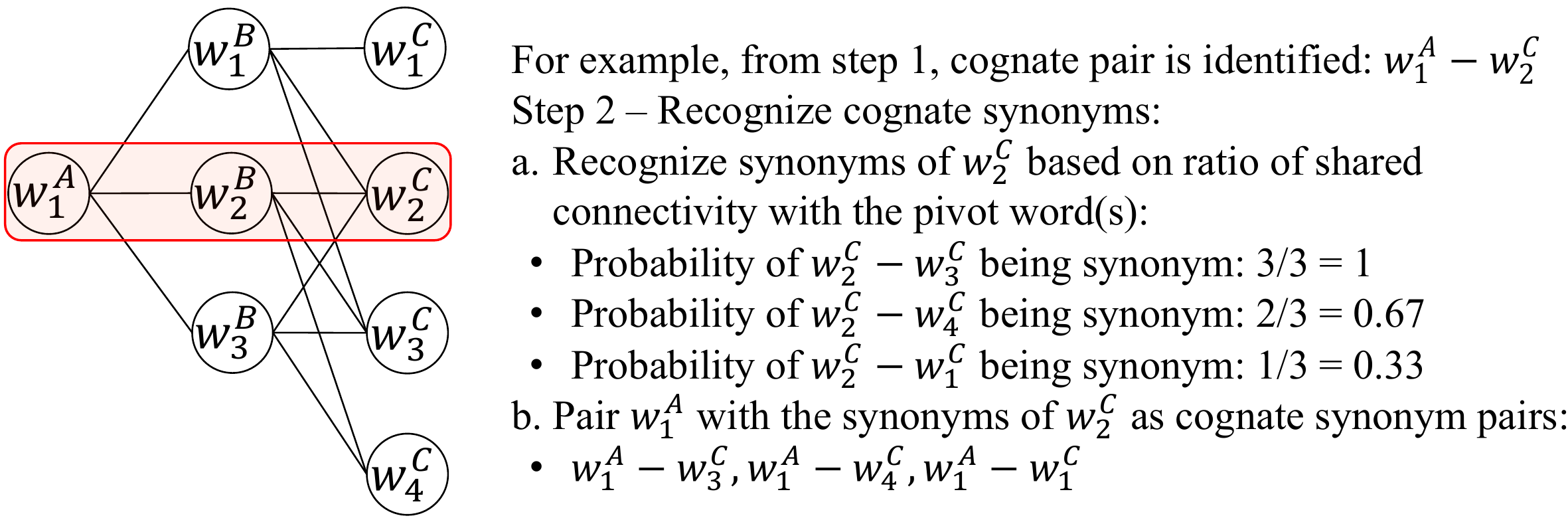} 
			\caption{Cognate Synonym Recognition.}
			\label{fig.4}
		\end{center}
	\end{figure}
	Finally, by recognizing both cognate pairs and cognate synonym pairs, we can obtain many-to-many translation results.
	\section{Generalization of Constraint-based Lexicon Induction Framework}
	We generalize the constraint-based lexicon induction framework by extending the existing one-cycle symmetry assumption into the n-cycle symmetry assumption and identify cognates and cognate synonyms by utilizing four heuristics to improve the quality and quantity of the translation pair results. 
	\subsection{Tripartite Transgraph}
	To model translation connectivity between language A and C via pivot language B, we define the tripartite transgraph, which is a tripartite graph in which a vertex represents a word and an edge represents the indication of shared meaning(s) between two vertices. Two tripartite transgraphs can be joined if there exists at least one edge connecting a pivot vertex in one tripartite transgraph to one non-pivot vertex in the other tripartite transgraph. To maintain the basic form of a tripartite transgraph with $n$ number of pivot words (at least 1 pivot per transgraph), each pivot word must be connected to at least one word in every non-pivot language, and there has to be a path connecting all pivot words via non-pivot words. Hereafter, we abbreviate the tripartite transgraph to transgraph. \par
	In this research, we assume that the input dictionaries contain no sense information. After modeling the translation connectivity from the input dictionaries as transgraphs, we further analyze the shared edges between the non-pivot vertices and the pivot vertices to predict the shared meanings between them. We then formalize the problem into Conjunctive Normal Form (CNF) and using WPMaxSAT solver to return the most probable correct translation results.\par
	Sometimes, for high-resource languages where the input dictionaries have many shared meanings via the pivot words, a big transgraph can be generated which potentially leads to excessive computational complexity when we formalize and solve it. Nevertheless, for low-resource languages where we can expect the input dictionaries to have just a few shared meanings via the pivot words, transgraph size is small enough to make its formalization and solution feasible. Therefore, for the sake of simplicity, we ignore big transgraphs in our experiments.
	\subsection{N-cycle Symmetry Assumption}
	Machine-readable bilingual dictionaries are rarely available for low-resource languages like Indonesian ethnic languages. It is even difficult to find sizable printed bilingual dictionary with acceptable quality for Indonesian ethnic languages. In the currently available machine-readable or printed dictionaries, we can expect to find missed senses/meanings that would lead to asymmetry in the transgraph. The expected missed senses are represented as dashed edges in the transgraph as depicted in Fig. 5(b). The one-to-one approach only considers translation pair candidates from existing connected solid edges in the transgraph as shown in Fig. 6(a). To fully satisfy symmetry constraint in the transgraph, we extend the existing one-cycle symmetry assumption to the n-cycle symmetry assumption while considering new translation pair candidates from the new dashed edges. As shown in Fig. 6(b), during the second cycle, the previously new dashed edges developed in the first cycle are taken to exist, therefore, we can extract translation pair candidates not only from the solid edges but also from the previously added dashed-edges. Users can input the maximum cycle for the experiment as shown in Algorithm 2 (as $maxCycle$).
	\begin{figure}[!h]
		\begin{center}
			\includegraphics[scale=1]{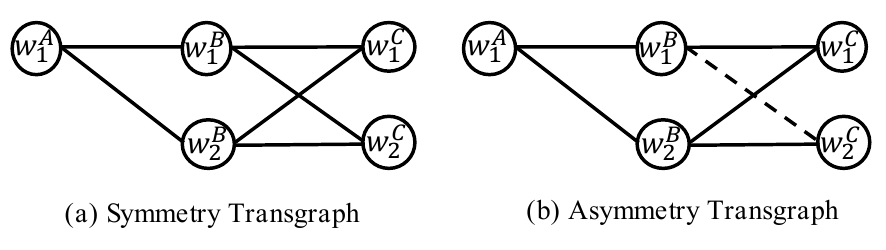} 
			\caption{Symmetry and Asymmetry Transgraphs.}
			\label{fig.5}
		\end{center}
	\end{figure}
	\begin{figure}[!h]
		\begin{center}																								
			\includegraphics[scale=0.6]{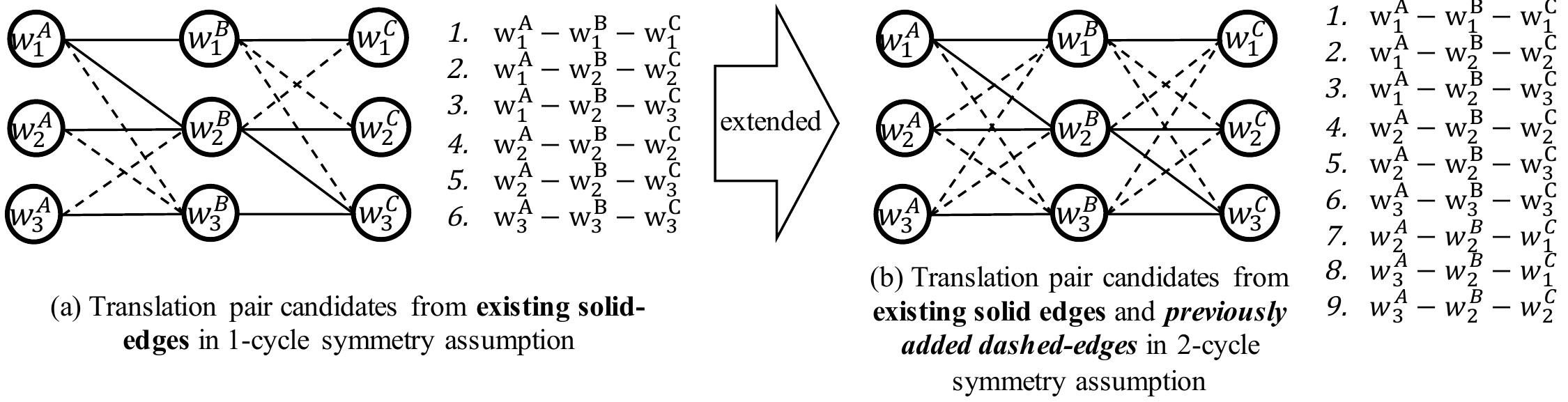} 
			\caption{N-cycle symmetry assumption extension.}
			\label{fig.6}
		\end{center}
	\end{figure}
	\subsection{Formalization}
	Constraint optimization problem formalism has been used in solving many natural language processing and web service composition related problems \cite{matsuno2011constraint,hassine2006constraint}. Our team \cite{wushouer-15} formalized bilingual lexicon induction as a WPMaxSAT problem. In this paper, we follow the same formulation. A literal is either a Boolean variable $x$ or its negation $\neg x$. A clause $C$ is a disjunction of literals $x_1 \vee ... \vee x_n$. A unit clause is a clause consisting of a single literal. A weighted clause is a pair ($C,\omega$), where $C$ is a clause and $\omega$ is a natural number representing the penalty for falsifying the clause $C$. If a clause is hard, the corresponding weight is infinity. 
	The propositional formula $\varphi_c^\omega$ in CNF \cite{biere2009handbook} is a conjunction of one or more clauses $C_1 \wedge ... \wedge C_n$. CNF formula with soft clauses is represented as $\varphi_c^+$ and $\varphi_c^\infty$ represents a CNF formula with hard clauses. The WPMaxSAT problem for a multiset of weighted clauses $C$ is the problem of finding an optimal assignment to the variables of $C$ that minimizes the cost of the assignment on $C$.
	Let $w_i^{A}$, $w_j^{C}$ and $w_k^{C}$ represents words from language $A, B$ and $C$. We define seven propositions as Boolean variables between a pair of words $w_i^A$, $w_j^B$ and $w_k^C$ as follows:
		\begin{itemize}
			\item{$e(w_i^A,w_j^B)$ and $e(w_j^B,w_k^C)$ represents edge existence between word pair from language A and B and from language B and C respectively,}
			\item{$c(w_i^{A},w_k^{C})$, $c(w_i^{A},w_n^{C})$ and $c(w_m^{A},w_k^{C})$ represents whether the word pair from language A and C is a cognate pair, and}	
			\item{$s(w_i^A,w_n^C)$ and $s(w_m^A,w_k^C)$ represents whether the word pair from language A and C is a cognate synonym pair}
		\end{itemize}
	
	To encode some of the constraints to CNF, we use a resolution approach based on the Boolean algebra rule of $p \rightarrow q \wedge r \Leftrightarrow (\neg p \vee q) \wedge (\neg p \vee r)$.
	In the framework, we define $E_E$ as a set of word pairs connected by existing edges, $E_N$ as a set of word pairs connected by new edges, $D_C$ as a set of translation pair candidates, $D_{Co}$ as a set of cognate pairs, $D_{NCo}$ as a set of non-cognate pairs, $D_{PCo}$ as a set of pivot words from language B which are connecting the current cognate pair, and $D_R$ as a set of all translation pair results returned by the WPMaxSAT solver.
	
	\subsection{Heuristics to Find Cognate}
	We define four heuristics to find cognates in the transgraph: cognate pair coexistence probability, missing contribution rate toward cognate pair coexistence, polysemy pivot ambiguity rate, and cognate form similarity. Based on our symmetry assumption, when $w_i^A$ and $w_k^C$ in a transgraph share all of their senses through pivot word(s) from language B, they are a potential cognate pair, where the cognate pair coexistence probability equals 1, the missing contribution equals 0 and the polysemy pivot ambiguity rate equals 0. When $w_i^A$ and $w_k^C$ have the same spelling, they are a potential cognate pair, where the cognate form similarity equals 1. Thus, when $w_i^A$ and $w_k^C$ are satisfying the symmetry assumption and also have the same spelling, we take them as the highest potential cognate pair in the transgraph.
	\subsubsection{Cognate Pair Coexistence Probability}
	Cognate pairs of language A and C are induced from two input bilingual dictionaries, i.e., Dictionary A-B and Dictionary B-C. We define two sets of event for Dictionary A-B ($w_i^A$ and $w_j^B$) where event $w_i^A$ represents connecting word $w_i^A$ of language A to words of language B represented by edges based on shared meaning between them. Similarly, event $w_j^B$ represents connecting word $w_j^B$ of language B to words of language A. We also define two sets of event for Dictionary B-C ($w_j^B$ and $w_k^C$) where event $w_j^B$ represents connecting word $w_j^B$ of language B to words of language C and event $w_k^C$ represents connecting word $w_k^C$ of language C to words of language B. A marginal probability $P(w_i^A)$ is a probability of $w_i^A$ connected to words of language B. A conditional probability $P(w_i^A|w_j^B)$ is a probability of $w_i^A$ connected to $w_j^B$ considering other words of language A that also connected to $w_j^B$. A joint probability $P(w_i^A,w_j^B)$ is a probability of $w_i^A$ interconnected to $w_j^B$. For example, in Fig. 7, $P(w_1^A) = 2/3$, since $w_1^A$ has two connected edges with words of language B out of 3 existing connected edges between words of language A and words of language B. The joint probability $P(w_1^A,w_1^B) = 1/3$, since any word from language A and any word from language B are only interconnected with 1 edge out of 3 existing connected edges between words of language A and words of language B. \par
	\begin{figure}[!h]
		\begin{center}
			\includegraphics[scale=0.8]{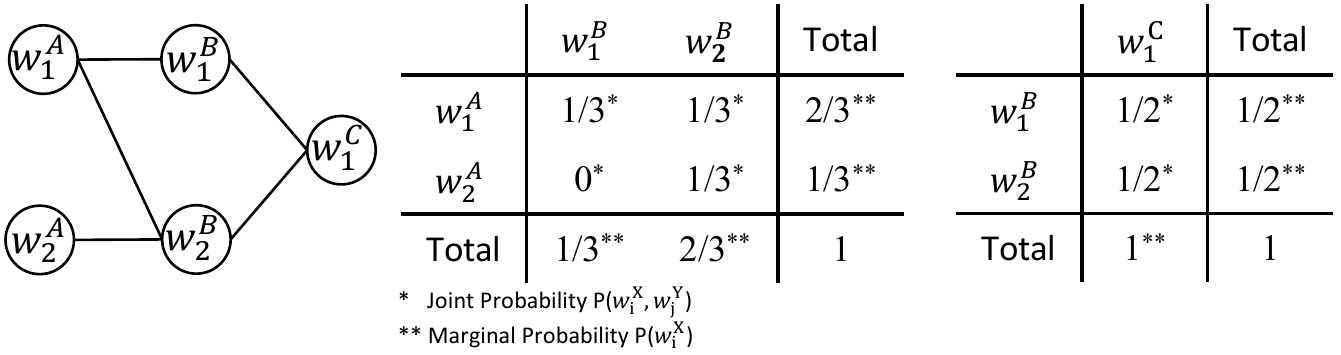} 
			\caption{Example of Marginal and Joint Probability.}
			\label{fig.7}
		\end{center}
	\end{figure}
	To calculate the possibility of a translation pair candidate $t(w_i^{A},w_k^{C})$ being a cognate pair $c(w_i^{A},w_k^{C})$, we calculate $t(w_i^A,w_k^C).H_{coex}$, a cognate coexistence probability of translation pair candidate $t(w_i^{A},w_k^{C})$. We firstly utilize a chain rule to obtain Eq.(1) and (2). By multiplying them, we can get Eq.(3). Event $w_i^A$ and event $w_k^C$ are independent since they are from a different input bilingual dictionary, thus, $P(w_k^C, w_i^A) = P(w_i^A)P(w_k^C)$ and Eq.(3) can be rewritten as Eq.(4). We use a generative probabilistic process which commonly used in prior work \cite{Dejean:2002,Wu2007,Nakov-12,richardson2015pivot} in Eq.(5) to obtain $P(w_i^A|w_k^C)$ and $P(w_k^C|w_i^A)$. Finally, we can obtain a cognate coexistence probability of translation pair candidate $t(w_i^{A},w_k^{C})$ as $t(w_i^A,w_k^C).H_{coex} = P(w_i^A, w_k^C)$.
	\begin{equation}
	\small
	P(w_i^A, w_k^C) = P(w_k^C|w_i^A)P(w_i^A)
	\end{equation}
	\begin{equation}
	\small
	P(w_k^C, w_i^A) = P(w_i^A|w_k^C)P(w_k^C)
	\end{equation}
	\begin{equation}
	\small
	P(w_i^A, w_k^C)P(w_k^C, w_i^A) = P(w_i^A|w_k^C)P(w_k^C|w_i^A)P(w_i^A)P(w_k^C)
	\end{equation}
	\begin{equation}
	\small
	P(w_i^A, w_k^C) = P(w_i^A|w_k^C)P(w_k^C|w_i^A)
	\end{equation}
	\begin{equation}
	\small
	P(w_i^A|w_k^C) = \sum_{j=0}P(w_i^A|w_j^B)P(w_j^B|w_k^C)
	\end{equation}
	\begin{figure}[!h]
		\begin{center}
			\includegraphics[scale=0.8]{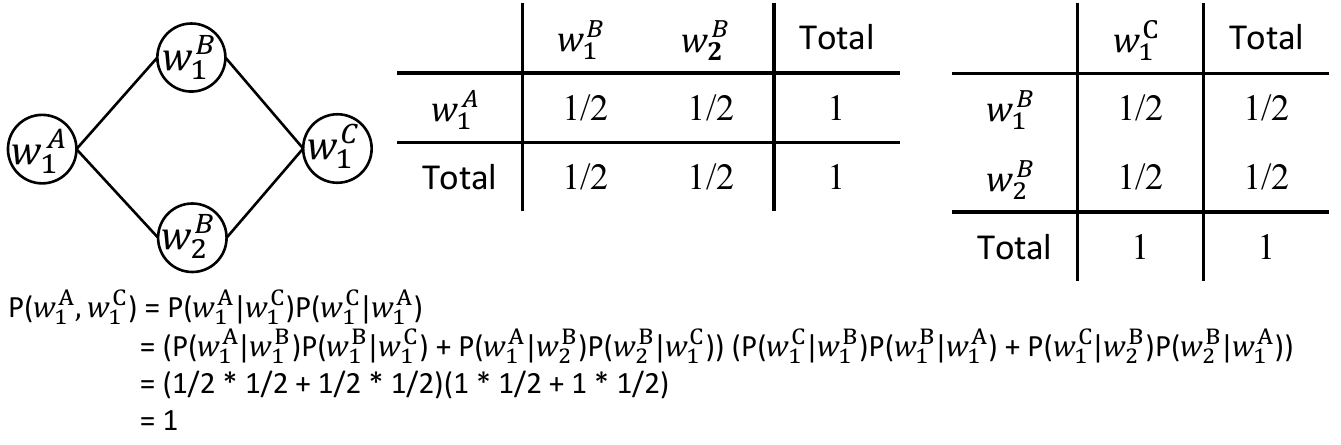} 
			\caption{Symmetry Pair Coexistence Probability.}
			\label{fig.8}
		\end{center}
	\end{figure}
	\begin{figure}[!h]
		\begin{center}
			\includegraphics[scale=0.8]{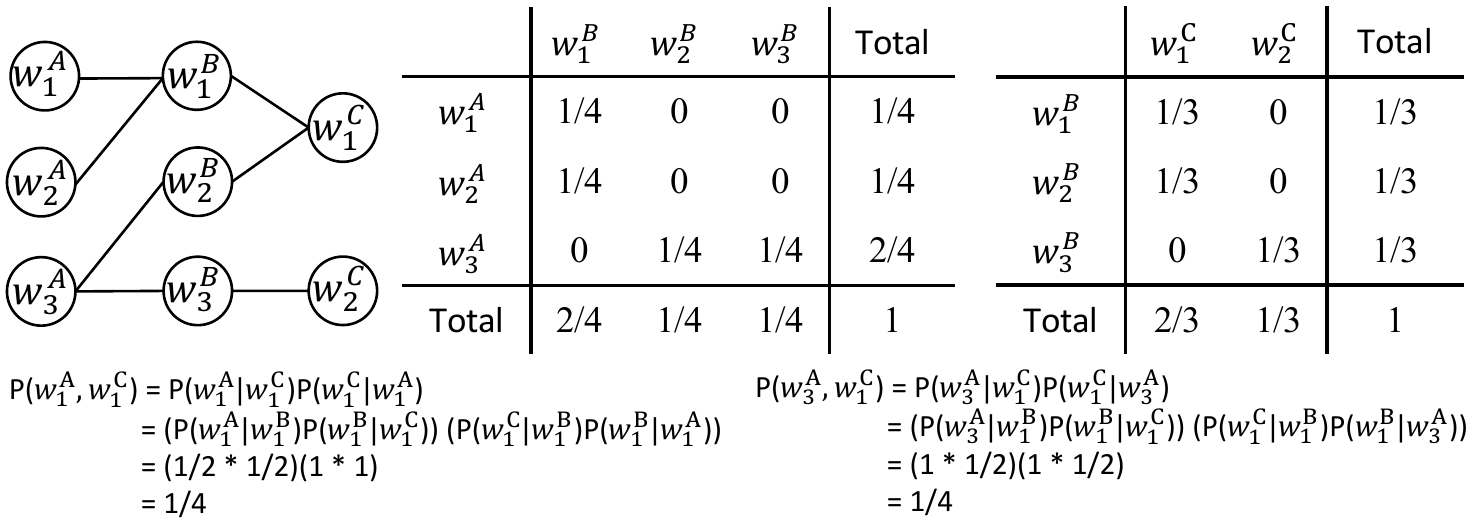} 
			\caption{Equal Treatment of Polysemy in Pivot/Non-Pivot Word.}
			\label{fig.9}
		\end{center}
	\end{figure}
	When $w_i^A$ and $w_k^C$ in a transgraph share all of their senses through pivot word(s) from language B and none of the pivot words are ambiguous, the cognate pair coexistence probability equals 1 as shown in Fig. 8. The algorithm to calculate the probability of the translation pair candidates coexisting as a cognate is shown in Algorithm 1 line number 19. The coexistence probability is very important in differentiating cognates from non cognates, but, it is poor at avoiding polysemy in pivot words. This is because it treats polysemy in the pivot words and polysemy in the non-pivot words equally. In reality, however, polysemy in pivot words negatively impacts the quality of bilingual dictionary induction rather than polysemy in non-pivot words. A case with high polysemy in pivot words and low polysemy in non-pivot words and a case with low polysemy in pivot words and high polysemy in non-pivot words where the two cases have equal rates of polysemy, will yield same probability as shown in Fig. 9. Therefore, we introduce a special heuristic to tackle this problem, i.e., polysemy pivot ambiguity rate.
	\subsubsection{Missing Contribution Rate Toward Cognate Pair Coexistence}
	Inspired by the Shapley Value \cite{shapley1953value}, a solution concept in cooperative game theory, we calculate missing contribution rate toward cognate pair coexistence probability by calculating coexistence probability of supposed cognate pair (also considering missing edges as existing) minus the coexistence probability of the pair from existing connectivity only. When $w_i^A$ and $w_k^C$ in a transgraph share all of their senses through pivot word(s) from language B (no missing senses), the missing contribution equals 0. The lower is the missing contribution toward coexistence probability of a translation pair candidate, the more likely is the translation pair candidate of being a cognate. The calculation of missing contribution rate of $w_1^A$ and $w_1^C$ pair, i.e., $t(w_i^A, w_k^C).H_{missCont}$ is shown in Algorithm 1 line number 20. \par
	\subsubsection{Polysemy Pivot Ambiguity Rate}
	To model the effect of polysemy in the pivot language on precision, for the sake of simplicity, we ignore synonym words within the same language. Polysemy in non-pivot languages have no negative effect on precision. In Fig. 10(a), even though the non-pivot words are connected by four pivot words representing four senses/meanings, the transgraph only has one translation pair candidate ($w_1^A$-$w_1^C$) and so the precision is 100\%.\par
	
	\begin{figure}[!h]
		\begin{center}
			\includegraphics[scale=0.8]{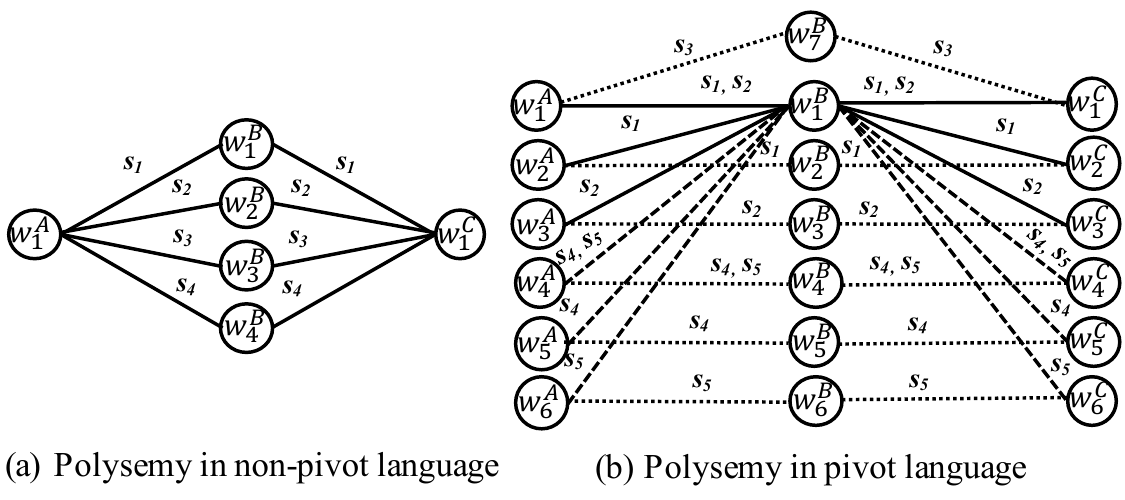} 
			\caption{Polysemy in pivot and non-pivot language.}
			\label{fig.10}
		\end{center}
	\end{figure}
	However, polysemy in pivot language negatively impacts precision. Fig. 10(b) shows that non-pivot word $w_1^A$ and $w_1^C$ are cognates and share the same meanings ($s_1$,$s_2$,$s_3$), but pivot word $w_1^B$ which has four meanings ($s_1$,$s_2$,$s_4$,$s_5$) only shares a part of the meanings ($s_1$,$s_2$) with the non-pivot words. The solid edges have part or all shared meanings ($s_1$, $s_2$) between the non-pivot words ($w_1^A$, $w_1^C$) and the pivot word $w_1^B$. The dashed edges express part or all unshared meanings ($s_4$, $s_5$) between the non-pivot words ($w_1^A$, $w_1^C$) and the pivot word $w_1^B$. To investigate the effect of pivot word $w_1^B$ on the overall precision, we extract only translation pair candidates from the connected edges. The precision (38.89\%) is affected negatively as there are 22 wrong translations because of the polysemy in pivot language ($w_1^B$) in the transgraph.\par
	We formalize the effect of polysemy in pivot language on precision with the following formulation where $n$ is the number of shared meanings between pivot word and non-pivot words and $m$ is the number of pivot meaning(s) that are not shared with non-pivot words. The number of correct translations contributed by the solid edges and the number of correct translations contributed by the dashed edges can be calculated by Eq.(6). The precision of the translation result is calculated by Eq.(7).\par
	\begin{equation}
	\small
	CorrectTrans(n) = 2\sum_{i=1}^{n}\sum_{j=1}^{i}\dbinom{n}{i}\dbinom{i}{j}-\sum_{i=1}^{n}\dbinom{n}{i}
	\end{equation}
	\begin{equation}
	\small
	Precision(n,m) = \frac{CorrectTrans(n) + CorrectTrans(m)}{\bigg[\sum_{i=1}^{n}\dbinom{n}{i} + \sum_{i=1}^{m}\dbinom{m}{i}\bigg]^2}
	\end{equation}\par
	We predict the effect of shared meanings between pivot word and non-pivot words by simulating ten sets of transgraphs with $n$ (the number of shared meanings between pivot word and non-pivot words) values ranging from 1 to 10 where, in each set, $m$ (the number of pivot meaning(s) that not shared with non-pivot words) ranges from 0 to $n$ in Fig. 11. In this experiment, non-pivot languages and pivot language are closely-related languages ($w_1^A$, $w_1^B$, and $w_1^C$ are cognates) when there is no pivot meaning that not shared with non-pivot words ($m=0$). This result shows that the greater the number of shared senses/meanings (represented by $n$) between pivot and non-pivot words there are, the lower the precision is. Nevertheless, the polysemy in the pivot language has the least negative effect on the precision when the pivot language and non-pivot languages are closely-related where the number of unshared pivot senses (represented by $m$) equals 0. The negative effect increases as the value of $m$ increases.\par
	\begin{figure}[!h]
		\begin{center}
			\includegraphics[scale=0.4]{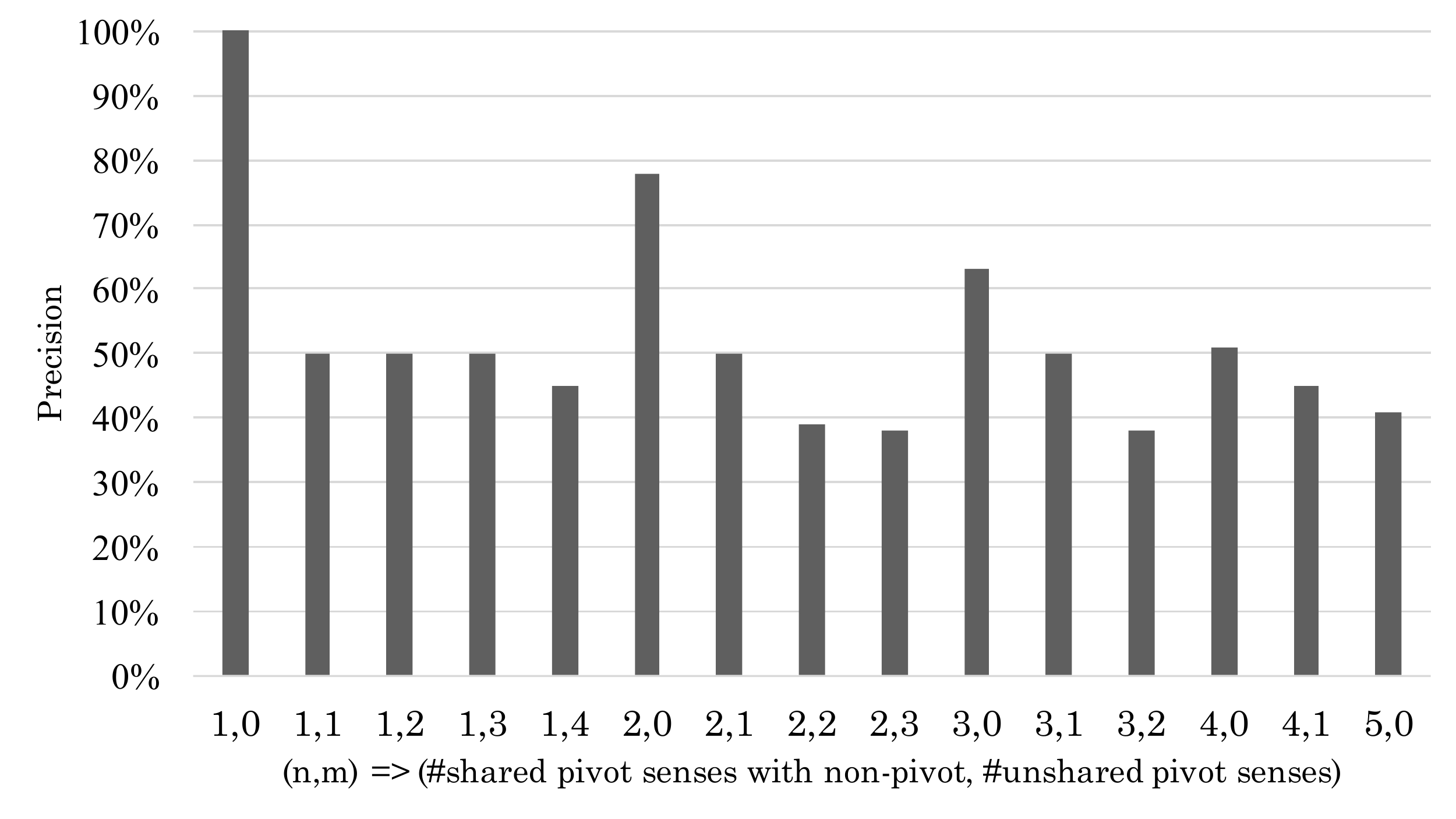} 
			\caption{Prediction model of precision on polysemy in pivot language.}
			\label{fig.11}
		\end{center}
	\end{figure}
	Polysemy in pivot words negatively impacts the precision of the translation result, unlike that in non-pivot words. Since we do not have any information about the senses from the input dictionaries, it is difficult to avoid the negative effect of the polysemous pivot word. To predict a probability of $w_i^A$ and $w_k^C$ to be a cognate pair via pivot word $w_j^B$ which share common senses, we assume the worst case scenario where the number of senses belonging to pivot word $w_j^B$ equals the maximum number of connected edges to $w_i^A$ or $w_k^C$. If the maximum number of indegree or outdegree of the polysemy pivot is $n$, there are $2^n-1$ possible combination of shared senses for every paths via pivot word $w_j^B$ in order for the translation pair candidates to be a cognate pair $c(w_i^A, w_k^C)$ out of all $(2^n-1)^2$ combinations. In Fig. 5(b), the possible combination of shared senses between $w_1^A$ and $w_1^C$ or between $w_1^A$ and $w_2^C$ are: [$s_1$, $s_2$, $s_1$ \& $s_2$]. To calculate the probability of the pair $w_i^A$ and $w_k^C$ being a cognate considering polysemy in the pivot words, we calculate $t(w_i^A, w_k^C).P_{sharedSenses}$, the product of the probabilities of shared senses between the pair for every existing path as shown in Algorithm 1 line number 10. The polysemy pivot ambiguity rate is given by $t(w_i^A, w_k^C).H_{polysemy} = 1 - t(w_i^A, w_k^C).P_{sharedSenses}$ as shown in Eq.(8) and Algorithm 1 line number 21.
	\begin{equation}
	\small
	t(w_i^A, w_k^C).H_{polysemy} = 1 - \prod \big((2^n-1)/(2^n-1)^2\big) = 1 - \prod \big(1 / (2^n-1)\big)
	\end{equation}\par
	The lower the polysemy pivot ambiguity rate is, the more likely it is that the pair form a cognate and share exact senses. When there is only one path between $w_i^A$ and $w_k^C$ and there is only one indegree and one outdegree of the pivot word $w_j^B$, the polysemy pivot ambiguity rate equals 0.
	
	\subsubsection{Cognate Form Similarity}
	Because the symmetry assumption can sometimes fail to select a cognate correctly when it gives the same cost for multiple translation pair candidates, the cognate form similarity heuristic will contribute to selecting the cognate. We calculate cognate form similarity using Longest Common Subsequent Ratio (LCSR) ranging from 0 (0\% form-similarity) to 1 (100\% form-similarity) following \cite{melamed1995induce} as shown in Eq.(9) and Algorithm 1 line number 22 where $LCS(w_i^A, w_k^C)$ is the longest common subsequence of $w_i^A$ and $w_k^C$; $|x|$ is the length of $x$; and $max(|w_i^A|, |w_k^C|)$ returns the longest length. However, the maximum cost contributed from the form dissimilarity is set at 1/100 of the maximum cost contributed by one symmetry assumption heuristic as shown in Algorithm 2 line number 24 to ensure that the cognate form similarity heuristic will have only a supporting role in helping the main symmetry assumption heuristics.
	
	\begin{equation}
	LCSR(w_i^A, w_k^C) = \frac{|LCS(w_i^A, w_k^C)|}{max(|w_i^A|, |w_k^C|)}
	\end{equation}
	\begin{equation}
	t(w_i^A, w_k^C).H_{formSim} = LCSR(w_i^A, w_k^C)
	\end{equation}
	
	\begin{algorithm*}
		\SetAlgoNoLine
		\DontPrintSemicolon
		\KwIn{Translation pair candidate $t(w_i^A,w_k^C)$;} 
		\KwOut{Translation pair candidate $t(w_i^A,w_k^C)$ with cognate pair probabilites information} 
		
		$P(w_i^A,w_k^C)$ = 0; $P(w_k^C,w_i^A)$ = 0; $P_{missing}(w_i^A,w_k^C)$ = 0; $P_{missing}(w_k^C,w_i^A)$ = 0;\;
		\For{each path in $t(w_i^A,w_k^C).Paths$}
		{
			$P(w_i^A|w_j^B)$ = 0; $P(w_j^B|w_k^C)$ = 0; $P(w_k^C|w_j^B)$ = 0; $P(w_j^B|w_i^A)$ = 0; \\
			\tcc{Conditional Probability direction: A-C}
			\lFor{each inEdge in $w_j^B.inEdges$ from language A}
			{
				$w_j^B.indegreeFromA \mathrel{+}= 1 / inEdge.Prob$;
			}
			\lFor{each inEdge in $w_k^C.inEdges$ from language B}
			{
				$w_k^C.indegreeFromB \mathrel{+}= 1 / inEdge.Prob$;
			}
			$P(w_i^A|w_j^B)$ = 1 / $w_j^B.indegreeFromA$; 
			$P(w_j^B|w_k^C)$ = 1 / $w_k^C.indegreeFromB$;\;		
			\tcc{Conditional Probability direction: C-A}
			\lFor{each inEdge in $w_j^B.inEdges$ from language C}
			{
				$w_j^B.indegreeFromC \mathrel{+}= 1 / inEdge.Prob$;
			}
			\lFor{each inEdge in $w_i^A.inEdges$ from language B}
			{
				$w_i^A.indegreeFromB \mathrel{+}= 1 / inEdge.Prob$;
			}
			$P(w_k^C|w_j^B)$ = 1 / $w_j^B.indegreeFromC$; 
			$P(w_j^B|w_i^A)$ = 1 / $w_i^A.indegreeFromB$;\; 
			
			$t(w_i^A,w_k^C).P_{sharedSenses} \mathrel{*}= 1 / (2^{max(w_j^B.indegreeFromA, w_j^B.indegreeFromC)} - 1)$;\;
			
			\eIf{missing edge exist in path}
			{$P_{missing}(w_i^A|w_k^C) \mathrel{+}= P(w_i^A|w_j^B) P(w_j^B|w_k^C)$;\;
				$P_{missing}(w_k^C|w_i^A) \mathrel{+}= P(w_k^C|w_j^B) P(w_j^B|w_i^A)$;\;}
			{$P(w_i^A|w_k^C) \mathrel{+}= P(w_i^A|w_j^B) P(w_j^B|w_k^C)$;\;
				$P(w_k^C|w_i^A) \mathrel{+}= P(w_k^C|w_j^B) P(w_j^B|w_i^A)$;}
		}
		$t(w_i^A,w_k^C).H_{coex} = P(w_i^A|w_k^C) P(w_k^C|w_i^A)$;\;
		$t(w_i^A,w_k^C).H_{missCont} = (P(w_i^A|w_k^C) + P_{missing}(w_i^A|w_k^C)) (P(w_k^C|w_i^A) + P_{missing}(w_k^C|w_i^A)) - (P(w_i^A|w_k^C) P(w_k^C|w_i^A))$;\;
		$t(w_i^A,w_k^C).H_{polysemy} = 1 - t(w_i^A,w_k^C).P_{sharedSenses}$;\;
	    $t(w_i^A,w_k^C).H_{formSim} = LCSR(w_i^A, w_k^C)$;\;
		return $t(w_i^A,w_k^C)$;
		
		\caption{Cognate Pair Probability Calculation}
		\label{alg:one}
	\end{algorithm*}
	
	\subsection{Constraints Extension}
	We extend the one-to-one approach constraints by adding several new constraints to the constraint sets to find cognates and cognate synonyms. All constraints are listed in Table 2.
	\subsubsection{Edge Existence}
	An edge exists in the transgraph between words that share their meaning(s) based on input dictionaries. The existing edges in the transgraph are encoded as TRUE, i.e., $e(w_i^{A},w_j^{B})$ and $e(w_j^{B},w_k^{C})$ in the CNF formula which is represented as hard constraint $\varphi_1^\infty$.
	\subsubsection{Edge Non-Existence}
	An edge does not exist in the transgraph between words that do not share their meaning(s) based on input dictionaries. We formalize the non-existence of edge in the transgraph by encoding the negation of the literal edge existence, i.e., $\neg e(w_i^{A},w_j^{B})$ and $\neg e(w_j^{B},w_k^{C})$ in the CNF formula which is represented as soft constraint $\varphi_2^+$.
	\subsubsection{Symmetry}
	Cognate share all of their senses / meanings and symmetrically connected via pivot language B. We convert $c(w_i^{A},w_k^{C}) \rightarrow e(w_i^{A},w_1^{B}) \wedge e(w_i^{A},w_2^{B}) \wedge ... \wedge e(w_1^{B},w_k^{C}) \wedge e(w_2^{B},w_k^{C}) \wedge ...$ into $(\neg c(w_i^{A},w_k^{C}) \vee e(w_i^{A},w_1^{B})) \wedge (\neg c(w_i^{A},w_k^{C}) \vee e(w_i^{A},w_2^{B})) \wedge .... \wedge (\neg c(w_i^{A},w_k^{C}) \vee e(w_1^{B},w_k^{C})) \wedge (\neg c(w_i^{A},w_k^{C}) \vee e(w_1^{B},w_k^{C})) \wedge ... $ It is encoded as hard constraint $\varphi_3^\infty$. Unfortunately, a problem arises with low-resource languages where the input dictionaries have no sense information and many senses are expected to be missed due to the small size of the dictionaries. To solve this problem, we add new edges so that cognate pairs share all of the meanings by violating the edge non-existence soft constraint $\varphi_2^+$ and paying a cost determined from user-selected heuristics (cognate pair coexistence probability, missing contribution rate toward the cognate pair coexistence probability, polysemy pivot ambiguity rate, and cognate form similarity). In other words, we assume the edges exist. The higher the cognate pair coexistence probability and the lower the missing contribution rate toward the cognate pair coexistence probability and the lower the polysemy pivot ambiguity rate and the higher the cognate form similarity, the more likely it is that the pair form a cognate, thus, the lower is the cost of adding any new edge to it, i.e., the new edge weight. The new edges in the transgraph is encoded as FALSE (NOT exist), i.e., $\neg e(w_i^{A},w_j^{B})$ or $\neg e(w_j^{B},w_k^{C})$ in the CNF formula and depicted as dashed edges in the transgraph. The weight of the new edge from non-pivot word $w_i^A$ to pivot word $w_j^B$ is defined as $\omega(w_i^{A},w_j^{B})$ and the weight of a new edge from pivot word $w_j^B$ to non-pivot word $w_k^C$ is defined as $\omega(w_j^{B},w_k^{C})$. Both of $\omega(w_i^{A},w_j^{B})$ and $\omega(w_j^{B},w_k^{C})$ values equal $t(w_i^{A},w_k^{C}).H_{coex} + t(w_i^{A},w_k^{C}).H_{missCont} + t(w_i^{A},w_k^{C}).H_{polysemy} + t(w_i^{A},w_k^{C}).H_{formSim}$ as shown in Algorithm 2 line number 21-24. 
	
	\begin{algorithm*}
		\SetAlgoNoLine
		\DontPrintSemicolon
		\KwIn{$transgraphs$, $maxCycle$, $threshold$, $HSelections$;} 
		\KwOut{$D_{Co}$ \tcc*{list of cognate pair results}} 
		\SetKwFunction{proc}{calculateEdgeCost}
		\SetKwProg{myproc}{Function}{}{}
		
		\For{each $transgraph$ in \proc{transgraphs})}
		{
			\tcc{Extract the most probable cognate pair and cognate synonym pair with total cost of violating constraints below the threshold iteratively}
			$CNF_{cognate} \leftarrow constructCNF_{cognate}(transgraph.D_C)$;\tcc*{following Eq.(11)}
			\While{$cognatePair \leftarrow SATSolver.solve(CNF_{cognate})$}{
				\If{$cognatePair.totalCost < cognateThreshold$}{$D_{Co}\leftarrow cognatePair$; $CNF_{cognate}.update()$;}	
			}
			$CNF_{cognateSynonym} \leftarrow constructCNF_{cognateSynonym}(transgraph.D_C)$;\tcc*{following Eq.(12)}
			\While{$cognateSynonymPair \leftarrow SATSolver.solve(CNF_{cognateSynonym})$}{
				\If{$cognateSynonymPair.totalCost < cognateSynonymThreshold$}{$D_{Co}\leftarrow cognateSynonymPair$; $CNF_{cognateSynonym}.update()$;}	
			}
		}
		return $D_{Co}$;
		
		\myproc{\proc{transgraphs}}{
			\For{each $transgraph$ in $transgraphs$}
			{
				$transgraph.D_C \leftarrow generateCandidates(transgraph)$; \tcc*{generate trans. pair cand.}
				\For{each $t(w_i^A,w_k^C)$ in $transgraph.D_C$}
				{
					calculateCognatePairProb($t(w_i^A,w_k^C)$);\tcc*{using Algorithm 1}
					\tcc{Cost of adding new edges are calculated from user selected heuristics}
					\lIf{HSelections.coex is TRUE}{ 
						$t(w_i^A,w_k^C).EdgeCost \mathrel{+}= 1 - t(w_i^A,w_k^C).H_{coex}$;
					}
					\lIf{HSelections.missCont is TRUE}{
						$t(w_i^A,w_k^C).EdgeCost \mathrel{+}= t(w_i^A,w_k^C).H_{missCont}$;
					}
					\lIf{HSelections.polysemy is TRUE}{
						$t(w_i^A,w_k^C).EdgeCost \mathrel{+}= t(w_i^A,w_k^C).H_{polysemy}$;
					}
					\lIf{HSelections.formSim is TRUE}{
						$t(w_i^A,w_k^C).EdgeCost \mathrel{+}= \big(1 - t(w_i^A,w_k^C).H_{formSim}\big)/100$;
					}
					\For{each $w_i^A.outEdges$}
					{
						\lIf{$e(w_j^B,w_k^C)$ is not exist}{$t(w_i^A,w_k^C).e(w_j^B,w_k^C).Cost = t(w_i^A,w_k^C).EdgeCost$;}
					}
					\For{each $w_i^C.inEdges$}
					{
						\lIf{$e(w_i^A,w_j^B)$ is exist}{$t(w_i^A,w_k^C).e(w_i^A,w_j^B).Cost = t(w_i^A,w_k^C).EdgeCost$;}
					}
				}
			}	
			\If{maxCycle is not reached}{
				$transgraphs \leftarrow addNewEdges()$;\tcc*{add new edges to transgraphs for the next cycle}
				\proc{transgraphs};
			}
			\nl \KwRet $transgraphs$\;}
		\caption{Cognate and Cognate Synonym Extraction}
		\label{alg:two}
	\end{algorithm*}
	\subsubsection{Uniqueness}
	The first step of our strategy in obtaining many-to-many translation pair with good quality is to extract a list of cognates in the transgraph. The uniqueness constraint ensures that only one-to-one cognates which share all of their meanings will be considered as translation pairs. In other words, a word in language A can only be a cognate with just one word from language C. This is encoded as hard constraint $\varphi_4^\infty$.
	\subsubsection{Extracting at Least One Cognate}
	Since the framework communicates with WPMaxSAT solver iteratively as shown in Algorithm 2 line number 2-7, hard constraint $\varphi_5^\infty$ ensures that at least one of the $c(w_i^A,w_k^C)$ variables must be evaluated as TRUE. Consequently, each iteration yields one most probable cognate pair and stores it in  $D_{Co}$ and also in $D_{R}$ as a translation pair result. This clause is a disjunction of all $c(w_i^A,w_k^C)$ variables.
	\subsubsection{Encoding Cognate}
	We exclude previously selected translation pairs, which are stored in $D_{Co}$ from the following list of cognate pair candidates by encoding them as TRUE, i.e., $c(w_i^A,w_k^C)$ which is encoded as hard constraint $\varphi_6^\infty$, and excluding them from $\varphi_5^\infty$.
	\subsubsection{Encoding Non-Cognate}
	Once we get list of all cognate pairs, stored in $D_{Co}$, the remaining translation pair candidates are stored in $D_{NCo}$ and encoded as FALSE, i.e., $\neg c(w_i^{A},w_k^{C})$ in the CNF formula, which is represented as hard constraint $\varphi_7^\infty$.
	
	\begin{table*}
		\caption{Constraints for Cognates and Cognate Synonyms Extraction}
		\footnotesize
		\label{tab:2}
		\begin{tabular}{ll}			
			\hline
			ID&CNF Formula\\
			\hline\hline
			&\textit{Edge Existence:} \\
			$\varphi_1^\infty$&$\Big(\bigwedge\limits_{(w_i^A,w_j^B) \in E_E}\big(e(w_i^{A},w_j^{B}),\infty\big)\Big)  \wedge \Big(\bigwedge\limits_{(w_j^B,w_k^C) \in E_E}\big(e(w_j^{B},w_k^{C}),\infty\big)\Big)$\\
			&\textit{Edge Non-Existence:}\\
			$\varphi_2^+$&$\Big(\bigwedge\limits_{(w_i^A,w_j^B) \in E_N}\big(\neg e(w_i^{A},w_j^{B}),\omega(w_i^{A},w_j^{B})\big)\Big) \wedge \Big(\bigwedge\limits_{(w_j^B,w_k^C) \in E_N}\big(\neg e(w_j^{B},w_k^{C}),\omega(w_j^{B},w_k^{C})\big)\Big)$\\
			&\textit{Symmetry:}\\
			$\varphi_3^\infty$&$\Big(\bigwedge\limits_{\substack{(w_i^A,w_j^B) \in E_E\cup E_N \\ (w_i^A,w_k^C) \in D_C}}\big((\neg c(w_i^A,w_k^C) \vee e(w_i^{A},w_j^{B})),\infty\big)\Big) \wedge \Big(\bigwedge\limits_{\substack{(w_j^B,w_k^C) \in E_E\cup E_N \\ (w_i^A,w_k^C) \in D_C}}\big((\neg c(w_i^A,w_k^C) \vee e(w_j^{B},w_k^{C})),\infty\big)\Big)$\\
			&\textit{Uniqueness:}\\
			$\varphi_4^\infty$&$\Big(\bigwedge \limits_{\substack{k \neq n \\ (w_i^A,w_k^C) \in D_C\\(w_i^A,w_n^C) \in D_C}}\big((\neg c(w_i^A,w_k^C) \vee \neg c(w_i^A,w_n^C)),\infty\big)\Big) \wedge \Big(\bigwedge \limits_{\substack{i \neq m \\ (w_i^A,w_k^C) \in D_C\\(w_i^A,w_n^C) \in D_C}}\big((\neg c(w_i^A,w_k^C) \vee \neg c(w_m^A,w_k^C)),\infty\big)\Big)$\\
			&\textit{Extracting at Least One Cognate:}\\
			$\varphi_5^\infty$&$\Bigg(\Big(\bigvee \limits_{(w_i^A,w_k^C) \notin D_R} c(w_i^A,w_k^C)\Big),\infty\Bigg) $\\
			&\textit{Encoding Cognate:}\\
			$\varphi_6^\infty$&$\bigwedge \limits_{(w_i^A,w_k^C) \in D_{Co}} \big(c(w_i^A,w_k^C),\infty\big)$\\
			&\textit{Encoding Non-Cognate:}\\
			$\varphi_7^\infty$&$\bigwedge \limits_{(w_i^A,w_k^C) \in D_{NCo}} \big(\neg c(w_i^A,w_k^C),\infty\big)$\\
			&\textit{Cognate Synonym:}\\
			$\varphi_8^\infty$&$\Bigg(\bigwedge \limits_{\substack{k \neq n \\ (w_i^A,w_k^C) \in D_{Co}\\(w_i^A,w_n^C) \notin D_R}}\big((\neg s(w_i^A,w_n^C) \vee  c(w_i^A,w_k^C)),\infty\big) \wedge \Big(\bigwedge \limits_{\substack{w_j^B \in D_{PCo}}}\big((\neg s(w_i^A,w_n^C) \vee  e(w_j^B,w_n^C)),\infty\big)\Big) \Bigg)$\\
			&$\wedge \Bigg(\bigwedge \limits_{\substack{i \neq m \\ (w_m^A,w_k^C) \in D_{Co} \\ (w_i^A,w_k^C) \notin D_R}}\big((\neg s(w_m^A,w_k^C) \vee  c(w_i^A,w_k^C)),\infty\big) \wedge \Big(\bigwedge \limits_{\substack{w_j^B \in D_{PCo}}}\big((\neg s(w_m^A,w_k^C) \vee  e(w_m^A,w_j^B)),\infty\big)\Big)\Bigg)$\\
			&\textit{Extracting at Least One Cognate Synonym:}\\
			$\varphi_9^\infty$&$\Bigg(\Big(\bigvee \limits_{(w_i^A,w_k^C) \notin D_R} s(w_i^A,w_k^C)\Big), \infty\Bigg) $\\
			
			\hline
		\end{tabular}
	\end{table*}
	
	\subsubsection{Cognate Synonym}
	We can further identify cognate synonyms to improve the quantity of the translation results. For each cognate pair $c(w_i^A, w_k^C)$  stored in $D_{Co}$, we can find cognate synonym pairs $s(w_i^A, w_n^C)$ and $s(w_m^A, w_k^C)$ by extracting synonyms of $w_k^C$ and $w_i^A$ respectively. We assume that synonymous words are connected to common pivot words. We can add new edges by paying cost of violating soft-constraint $\varphi_2^+$ with a weight different from that used when identifying cognate pairs in the first step. In this second step, the weight is calculated based on cognate synonym probability $P_{cognateSyn}$ for both $w_n^C - w_k^C$ and $w_m^A - w_i^A$ based on percentage of shared connectivity with the pivot words. The weight, i.e., $1 - P_{cognateSyn}$ is distributed evenly to each new edges. We convert $s(w_i^{A},w_n^{C}) \rightarrow c(w_i^{A},w_k^{C}) \wedge e(w_1^{B},w_n^{C}) \wedge e(w_2^{B},w_n^{C}) \wedge ... $ into $(\neg s(w_i^{A},w_n^{C}) \vee c(w_i^{A},w_k^{C})) \wedge (\neg s(w_i^{A},w_n^{C}) \vee e(w_1^{B},w_n^{C})) \wedge (\neg s(w_i^{A},w_n^{C}) \vee e(w_2^{B},w_n^{C})) \wedge ....$ With the same rule, we convert $s(w_m^{A},w_k^{C}) \rightarrow c(w_i^{A},w_k^{C}) \wedge e(w_m^{A},w_1^{B}) \wedge e(w_m^{A},w_2^{B}) \wedge ... $ into $(\neg s(w_m^{A},w_k^{C}) \vee c(w_i^{A},w_k^{C})) \wedge (\neg s(w_m^{A},w_k^{C}) \vee e(w_m^{A},w_1^{B})) \wedge (\neg s(w_m^{A},w_k^{C}) \vee e(w_m^{A},w_2^{B})) \wedge ....$ It is encoded as hard constraint $\varphi_8^\infty$. In Fig. 4, $s(w_1^A, w_3^C).P_{cognateSyn} = 1$, $s(w_1^A, w_4^C).P_{cognateSyn} = 0.67$, and $s(w_1^A, w_1^C).P_{cognateSyn} = 0.33.$ Another example, in Fig. 5(a), if cognate pair $c(w_1^A, w_1^C)$ is identified, we need to identify cognate synonym probability of $w_1^A$ (no candidate exist) and $w_1^C$ (candidate: $w_2^C$). Based on the rate of shared connectivity with pivot word(s), $s(w_1^A, w_2^C).P_{cognateSyn} = 2/2$ and in Fig. 5(b) with the same way we can get $s(w_1^A, w_2^C).P_{cognateSyn} = 1/2$. 
	\subsubsection{Extracting at Least One Cognate Synonym}
	In the second step, i.e., finding cognate synonyms, the framework also communicates with the WPMaxSAT solver iteratively as shown in Algorithm 2 line number 8-13, and hard constraint $\varphi_9^\infty$ ensures that at least one of the $s(w_i^A,w_n^C)$ variables or $s(w_m^A,w_k^C)$ variables must be evaluated as TRUE. Consequently, each iteration yields one most probable cognate synonym pair and store it in $D_{R}$ as a translation pair result. This clause is a disjunction of all $s(w_i^A,w_k^C)$ variables.
	
	\subsection{Framework Generalization}
	We define two main CNF formulas; one for recognizing cognate pairs, i.e., $CNF_{cognate}$ as shown in Eq.(11) and one for recognizing cognate synonym pairs, i.e., $CNF_{cognateSynonym}$ as shown in Eq.(12). We also define another CNF formula, i.e., $CNF_{M-M}$ as shown in Eq.(13) which extract many-to-many translation pairs by ignoring uniqueness constraint of the one-to-one approach \cite{NASUTION16.1238}. Three constraints are shared by the CNF formulas: $\varphi_{1}^\infty$, $\varphi_{2}^+$ and $\varphi_{6}^\infty$. 
	\begin{equation}
	CNF_{cognate} = \varphi_{1}^\infty \wedge \varphi_{2}^+ \wedge \varphi_{3}^\infty \wedge \varphi_{4}^\infty \wedge \varphi_{5}^\infty \wedge \varphi_{6}^\infty
	\end{equation}
	\begin{equation}
	CNF_{cognateSynonym} = \varphi_{1}^\infty \wedge \varphi_{2}^+ \wedge \varphi_{6}^\infty \wedge \varphi_{7}^\infty \wedge \varphi_{8}^\infty \wedge \varphi_9^\infty
	\end{equation}
	\begin{equation}
	CNF_{M-M} = \varphi_{1}^\infty \wedge \varphi_{2}^+ \wedge \varphi_{3}^\infty \wedge \varphi_{5}^\infty \wedge \varphi_{6}^\infty
	\end{equation}
	\par
	Various constraint-based bilingual dictionary induction methods can be constructed to suit different situations and purposes by using a cognate recognition ($CNF_{cognate}$) or a cognate \& cognate synonym recognition ($CNF_{cognate} + CNF_{cognateSynonym}$) methods with a choice of n-cycle symmetry assumption, and with a series of individual and combined heuristics to be chosen as shown in Table 3. We can also define many-to-many translation pair extraction method in our previous work using $CNF_{M-M}$. Thus, we define our methods using Backus Normal Form as follow:\\
	$\langle situatedMethod \rangle ::= \langle cycle \rangle ":" \langle method \rangle ":" \langle heuristic\rangle$\\
	$\langle cycle \rangle ::= "1" | "2" | "3" | "4" | "5" | "6" | "7" | "8" | "9"$\\
	$\langle method \rangle ::= "C" | "S" | "M"$\\
	$\langle heuristic \rangle ::= "H1" | "H2" | "H3" | "H4" | "H12" | "H13" | "H14" | "H23" | "H24"|"H123"|"H124"|"H234"$
	\begin{itemize}
		\item {\textit{cycle}: symmetry assumption cycle where cycle $\geq$ 1.}
		\item {\textit{method}: \textit{C} as a cognate recognition ($CNF_{cognate}$) or \textit{S} as a cognate \& cognate synonym recognition ($CNF_{cognate} + CNF_{cognateSynonym}$) or \textit{M} as a many-to-many approach ($\Omega_2$ \& $\Omega_3)$ in our previous work \cite{NASUTION16.1238}.}
		\item {\textit{heuristic}: an individual or combined heuristics where H1234 means a combination of heuristic 1 (cognate pair coexistence probability), heuristic 2 (missing contribution rate toward cognate pair coexistence), heuristic 3 (polysemy pivot ambiguity rate), and heuristic 4 (cognate form similarity).}
	\end{itemize}
	A combination of cognate only ($CNF_{cognate}$) method with 1-cycle symmetry assumption and heuristic 1 is defined as 1:C:H1, yielding an identical method with one-to-one approach \cite{Wushouer2014} and $\Omega_1$ in our prior work \cite{NASUTION16.1238}. A combination of cognate only ($CNF_{M-M}$) method with heuristic 1 and 1-cycle symmetry assumption is defined as 1:M:H1, 
	which is identical with $\Omega_2$ 
	and for 2-cycle symmetry assumption is defined as 2:M:H1, which is identical with $\Omega_3$ in our prior work \cite{NASUTION16.1238}.
	\begin{table}[h!]
		\caption{Variation of Constraint-based Bilingual Dictionary Induction}
		\small
		\label{tab:3}
		\begin{threeparttable}
			\begin{center}
				\begin{tabular}{l|r|r|r}
					\toprule
					Cycle&$CNF_{cognate}$&$CNF_{cognate} + CNF_{cognateSynonym}$&$CNF_{M-M}$\\
					\hline
					1&H1\tnote{1}, H2, H3, H4, H12, ...&H1, H2, H3, H4, H12, ...&H1\tnote{2}\\
					
					\textgreater1&H1, H2, H3, H4, H12, ...&H1, H2, H3, H4, H12, ...&H1\tnote{3}\\
					\bottomrule
				\end{tabular}
				\begin{tablenotes}
					\item[1] \textit{Identical to one-to-one approach \cite{Wushouer2014} and $\Omega_1$ in our prior work \cite{NASUTION16.1238}}			
					\item[2] \textit{Identical to $\Omega_2$ in our prior work \cite{NASUTION16.1238}}
					\item[3] \textit{For 2-cycle, identical to $\Omega_3$ in our prior work \cite{NASUTION16.1238}}
				\end{tablenotes}
			\end{center}
		\end{threeparttable}
	\end{table}

	\section{Experiment}
	To evaluate our result, we calculate precision, recall and the harmonic mean of precision and recall using the traditional F-measure or balanced F-score \cite{Rijsbergen:1979:IR:539927}. In each iteration, WPMaxSAT solver returns the optimal translation pair result with minimum total cost (incurred by violating some soft constraints). Translation pair result with total cost above the threshold are not considered. For the methods equivalent with our prior work \cite{NASUTION16.1238} which are 1:C:H1, 1:M:H1, and 2:M:H1, we do not set any threshold. We try to analyze the impact of the threshold and the heuristics on the precision, recall and F-score. For this purpose, we need to have a Gold Standard, so that for each experiment, we can iterate threshold from 0 to the highest cost of constraint violation cost with 0.01 interval and try every combination of heuristics as input to Algorithm 2 (as $threshold$ \& $HSelections$) while observing the resulting precision, recall or F-score after evaluation against the gold standard. In this paper, we choose the result with the highest F-score. We want to analyze the algorithm so that our generalized constraint approach can be applied to other datasets for various languages. We conduct experiments with 6 methods constructed from our generalized constraint approach in which 3 of them yielding one-to-one translation pairs (1-1), i.e., Cognates recognition with all combination of heuristic and 1-cycle symmetry assumption (1:C:$\langle heuristic\rangle$), 2-cycles symmetry assumption (2:C:$\langle heuristic\rangle$), and 3-cycles symmetry assumption (3:C:$\langle heuristic\rangle$), and the rest yielding many-to-many translation pairs (M-M), i.e., Cognate and Cognate Synonyms recognition with all combination of heuristic and 1-cycle symmetry assumption (1:S:$\langle heuristic\rangle$), 2-cycles symmetry assumption (2:S:$\langle heuristic\rangle$), and 3-cycles symmetry assumption (3:S:$\langle heuristic\rangle$). As baselines, we use three methods from our previous work where H1 is the sole heuristic used \cite{NASUTION16.1238}, i.e., one-to-one translation pair extraction ($\Omega_1$) which is defined as 1:C:H1, many-to-many translation pair extraction from connected existing edges ($\Omega_2$) which is defined as 1:M:H1, and many-to-many translation pair extraction from connected existing and new edges ($\Omega_3$) which is defined as 2:M:H1. We also use the inverse consultation method (IC) and translation pairs generated from Cartesian product of input dictionaries (CP) as baselines.
	
	\subsection{Experimental Settings}
	We have four case studies; one of the closely related low-resource languages of Austronesian language family and three of high-resource Indo-European languages. The language similarities shown in Table 4 were computed using ASJP. We generate translation pairs from Cartesian Product within and across transgraph to be used in the evaluation as shown in Fig. 12.\par
	\begin{table}
		\caption{Language Similarity of Input Dictionaries}
		\label{tab:4}
		\begin{center}
			\begin{tabular}{lrrr}
				\toprule
				Language Pair&Language Similarity\\
				\hline
				min-ind, zlm-ind, min-zlm &69.14\%, 87.70\%, 61.66\%\\
				deu-eng, nld-eng, deu-nld &31.38\%, 39.27\%\, 51.17\%\\
				spa-eng, por-eng, spa-por&6.66\%, 3.79\%, 32.04\%\\
				deu-eng, ita-eng, deu-ita &31.38\%, 9.75\%, 13.64\%\\
				\bottomrule
			\end{tabular}
		\end{center}
	\end{table}
	\begin{figure}[!h]
		\begin{center}
			\includegraphics[scale=0.7]{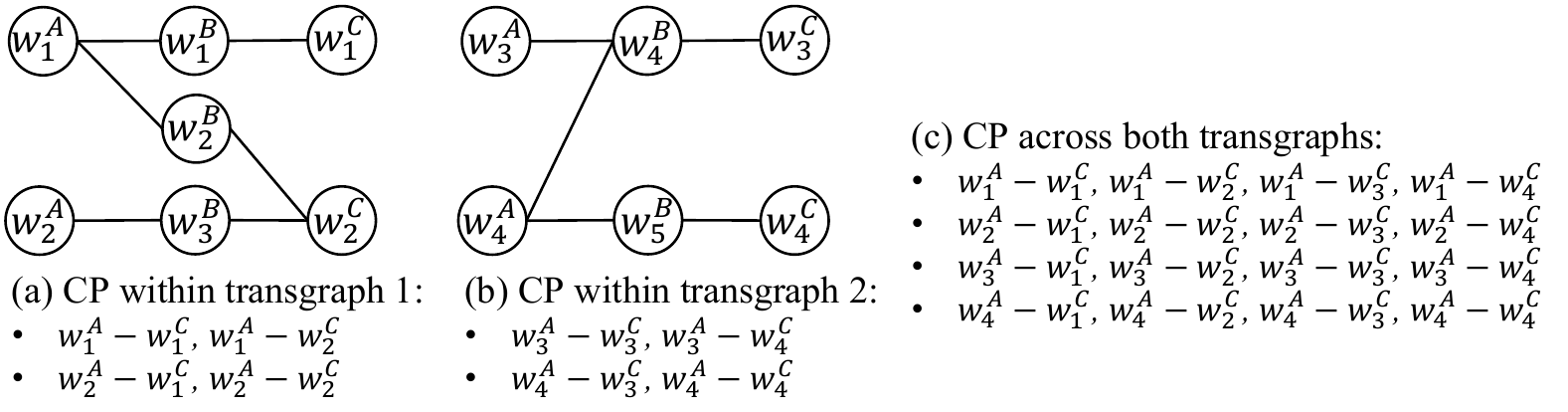} 
			\caption{Example of Extracting Translation Pair Candidates from Cartesian Product (CP).}
			\label{fig.12}
		\end{center}
	\end{figure}
	We selected Indonesian ethnic languages Minangkabau (min) and Riau Mainland Malay (zlm) with Indonesian language (ind) as the pivot for our first case study (min-ind-zlm). Even though Malaysian Malay (zlm) is not part of Indonesian ethnic languages, but it is very similar with Riau Mainland Malay. In fact, Riau Mainland Malay is one of Malaysian Malay dialects \cite{ethnologue-17}. Since there is no available machine readable dictionary of Indonesian to Riau Mainland Malay, we used the available machine readable dictionary of Indonesian to Malaysian Malay (zlm) for case study min-ind-zlm. A trilingual Indonesian, Malaysian Malay and Riau Mainland Malay speaker thoroughly cleansed the dictionary by deleting or editing Malaysian Malay words that are not present in the Riau Mainland Malay language. We generate full-matching translation pairs (Cartesian product within transgraph from input dictionaries), verified by the Minangkabau-Malay bilingual speaker via crowdsourcing and took them as the gold standard for calculating precision and recall.\par
	The Proto-Indo-European language is the common ancestor of the Indo-European language family from which the rest of our case study languages originate. The second case study (deu-eng-nld) targets high-resource languages of German (deu) and Dutch (nld) with English (eng) as the pivot. The third case study (spa-eng-por) uses Spanish (spa) and Portugese (por) languages with English (eng) as the pivot. The fourth case study (deu-eng-ita) uses German (deu) and Italian (ita) languages with English (eng) as the pivot. We utilize Freedict, an open source online bilingual dictionary databases\footnote{http://freedict.org} as input dictionaries and combination of Freedict, Panlex - another bilingual dictionary databases\footnote{http://panlex.org}, and Google Translate\footnote{http://translate.google.com} as shown in Table 5 as dictionaries for evaluation to create a gold standard. We use Google Translate to translate all headwords from Cartesian Product (CP) within the transgraphs. The gold standard is obtained by intersecting the combination of dictionaries for evaluation with CP across transgraph as shown in Fig. 13. The structure of the input dictionaries and the gold standard for every case studies can be found in Table 6. The translation relationship of the input dictionaries varies from one-to-one until one-to-eight as shown in Table 7. For the low-resource case study, i.e., min-ind-zlm, the input dictionaries only have few one-to-many translation relations compared to the high-resource case studies. This shows that there are many potential missing senses in the input dictionaries. Consequently, sometimes we can miss some translation pair candidates across the transgraphs. Therefore, in this paper, we limit our scope to extracting translation pairs within the transgraphs.\par
	\begin{table}
		\caption{Dictionaries for Evaluation}
		\small
		\label{tab:5}
		\begin{threeparttable}
			\begin{center}
				\begin{tabular}{lrrr}
					\toprule
					Source&\multicolumn{3}{c}{Number of Translation}\\
					\hline
					Freedict&deu-nld $\cup$ nld-deu = 35,962&spa-por = 333&deu-ita $\cup$ ita-deu = 6,152\\
					Panlex&deu-nld = 405,076&spa-por = 343,665&deu-ita = 475,461\\
					Google Translate\tnote{*}&deu-nld $\cup$ nld-deu = 1,924&spa-por $\cup$ por-spa = 1,338&deu-ita $\cup$ ita-deu = 1,790\\
					\hline
					TOTAL&deu-nld = 406,370&spa-por = 344,126&deu-ita = 476,172\\
					\bottomrule
				\end{tabular}
				\begin{tablenotes}
					\item[*] \textit{Translating all headwords from CP within the transgraphs.}			
				\end{tablenotes}
			\end{center}
		\end{threeparttable}
	\end{table}
	We do not discriminate both single-word and multi-words expressions in the input dictionaries. After constructing the transgraphs from the input dictionaries, we find one big transgraph for each high-resource language case study as shown in Table 8. Sometimes, for high-resource languages where the input dictionaries have many shared meanings via the pivot words, a big transgraph can be generated which potentially leads to a computational complexity when we formalize and solve it. Nevertheless, for a low-resource languages where we can expect the input dictionaries only have a few shared meanings via the pivot words, the size of the transgraph is feasible to be formalized and solved. Therefore, for the sake of simplicity, we ignore any big transgraphs in these experiments.\par
	Different users are likely to have different motivation, priority and preference when creating a bilingual dictionary. For high-resource languages, some users tend to priorities precision over recall while for low-resource languages, most users tend to priorities recall to enrich the language resource. In this paper, we optimize the hyperparameters (cognate threshold and cognate synonym threshold) with a grid search by incrementing the cognate threshold from 0 to the highest cost of violating the constraints with 0.01 intervals and incrementing the cognate synonym threshold from 0 to 1 with 0.01 intervals in order to find the highest F-score.
	\par
	\begin{figure}[!h]
		\begin{center}
			\includegraphics[scale=0.2]{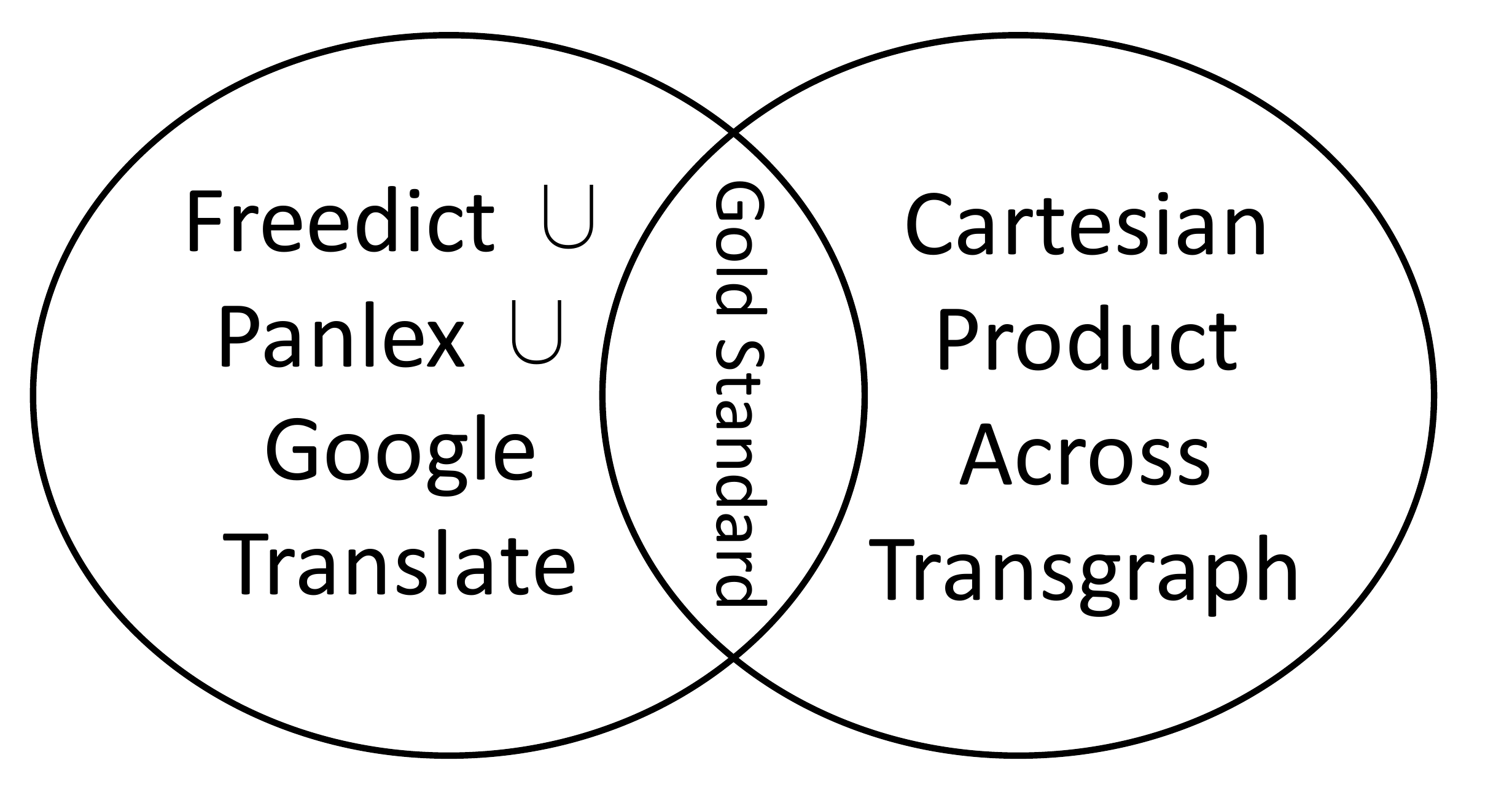} 
			\caption{Creating Gold Standard for the High-Resource Case Studies.}
			\label{fig.13}
		\end{center}
	\end{figure}
	\begin{table}
		\caption{Structure of Input Dictionaries and Gold Standard}
		\small
		\label{tab:6}
		\begin{center}
			\begin{tabular}{l|rrr|rrr|rrr|rrr}
				\toprule
				Case Study&\multicolumn{3}{c|}{min-ind-zlm}&\multicolumn{3}{c|}{deu-eng-nld}&\multicolumn{3}{c|}{spa-eng-por}&\multicolumn{3}{c}{deu-eng-ita}\\
				\hline
				Language&min&ind&zlm&deu&eng&nld&spa&eng&por&deu&eng&ita\\
				Headword&520&625&681&968&673&1,183&600&849&986&1,157&1,340&842\\
				\hline
				CP within transgraph&\multicolumn{3}{r|}{1,757}&\multicolumn{3}{r|}{5,790}&\multicolumn{3}{r|}{2,526}&\multicolumn{3}{r}{2,959}\\
				CP across transgraph&\multicolumn{3}{r|}{354,120}&\multicolumn{3}{r|}{1,145,144}&\multicolumn{3}{r|}{591,600}&\multicolumn{3}{r}{974,194}\\
				Gold Standard&\multicolumn{3}{r|}{1,246}&\multicolumn{3}{r|}{1,438}&\multicolumn{3}{r|}{1,069}&\multicolumn{3}{r}{1,503}\\
				\bottomrule
			\end{tabular}
		\end{center}
	\end{table}
	\begin{table}
		\caption{Translation Relationship of Input Dictionaries}
		\small
		\label{tab:7}
		\begin{center}
			\begin{tabular}{llrrrrrrrrrr}
				\toprule
				\multirow{2}{*}{Case Study}&\multirow{2}{*}{Bilingual Dictionary}&\multicolumn{8}{c}{Translation Relationship}\\
				{}&{}&1-1&1-2&1-3&1-4&1-5&1-6&1-7&1-8\\
				\hline
				\multirow{2}{*}{min-ind-zlm}
				{}&min-ind&267&210&36&5&1&1&0&0\\
				{}&zlm-ind&563&115&3&0&0&0&0&0\\
				\hline
				\multirow{2}{*}{deu-eng-nld}
				{}&deu-eng&785&165&16&2&0&0&0&0\\
				{}&nld-eng&705&410&49&14&3&1&1&0\\
				\hline
				\multirow{2}{*}{spa-eng-por}
				{}&spa-eng&204&289&86&16&2&2&1&0\\
				{}&por-eng&458&370&116&33&7&2&0&0\\
				\hline
				\multirow{2}{*}{deu-eng-ita}
				{}&deu-eng&971&154&30&2&0&0&0&0\\
				{}&ita-eng&256&421&129&25&7&2&1&1\\
				\bottomrule
			\end{tabular}
		\end{center}
	\end{table}
	\begin{table}
		\caption{Size of the Biggest Transgraph}
		\small
		\label{tab:8}
		\begin{center}
			\begin{tabular}{lrrrr}
				\toprule
				Case Study ($L_1-P-L_2$)&$L_1$ Words&$P$ Words&$L_2$ Words&Edges\\
				min-ind-zlm&8&14&18&39\\
				deu-eng-nld&4,669&2,486&6,864&18,548\\
				spa-eng-por&2,347&2,465&4,460&15,043\\
				deu-eng-ita&650&822&597&2,242\\
				\bottomrule
			\end{tabular}
		\end{center}
	\end{table}
	
	\subsection{Experiment Result}
	In all experiments and all case studies, all transgraphs are fully symmetrically connected on the third cycle, thus all possible translation pair candidates are reached. To extract many-to-many translation pairs, in the first step, i.e., cognate recognition and the second step, i.e., cognate synonym recognition, the soft-constraint violation threshold is set to reject all translation pairs returned by SATSolver that incurred a higher cost than the cognate threshold and cognate synonym threshold as shown in Algorithm 2 line number 4 and 10, respectively. Even though using the threshold to prioritize precision could yield the highest precision, the recall can be very low. Similarly, even though using the threshold to prioritize recall could yield the highest recall, the precision can also be harmed. Blindly prioritizing the precision over the recall or recall over the precision might not be a good strategy when implementing the framework.\par
	\subsubsection{Threshold Yielding The Highest F-score}
	\begin{table*}[!h]
		\caption{Threshold Yielding The Highest F-score}
		\footnotesize
		\label{tab:11}
		\begin{threeparttable}
			\begin{center}								
				\begin{tabular}{clrrrrrr}
					\toprule
					Case Study&Method&Cognate Threshold&Cognate Synonym Threshold&Precision&Recall&F-score&\\
					\hline
					\multirow{11}{*}{min-ind-zlm}
					{}&3:S:H14 (M-M) &4.79&1& 0.656 & 0.998 & 0.792\\
					{}&2:S:H14 (M-M) &4.79&0.74& 0.735 & 0.923 & 0.818\\
					{}&1:S:H14 (M-M) &4.17&1& 0.836 & 0.713 & 0.770\\
					{}&3:C:H14 (1-1) &4.79&&0.884 & 0.331 & 0.481\\
					{}&2:C:H14 (1-1) &4.79&&0.884 & 0.331 & 0.481\\
					{}&1:C:H14 (1-1) &4.17&&0.878 & 0.328 & 0.478\\
					{}&Baseline: 2:M:H1 (M-M)&&& 0.713 & 0.953 & 0.815\\
					{}&Baseline: 1:M:H1 (M-M)&&& 0.836 & 0.713 & 0.770\\
					{}&Baseline: 1:C:H1 (1-1)&&& 0.873 & 0.327 & 0.475\\
					{}&Baseline: CP (M-M)&&& 0.654 & 0.998 & 0.791\\
					{}&Baseline: IC (M-M) &&&0.950 & 0.031 & 0.059\\
					\hline
					\multirow{11}{*}{deu-eng-nld}
					{}&3:S:H14 (M-M) &1.97&1& 0.230 & 0.926 & 0.368\\
					{}&2:S:H14 (M-M) &1.97&0.49& 0.323 & 0.707 & 0.443\\
					{}&1:S:H124 (M-M) &4.1&0.99& 0.400 & 0.820 & 0.537\\
					{}&3:C:H14 (1-1)&1.97&&0.474 & 0.250 & 0.328\\
					{}&2:C:H14 (1-1)&1.97&&0.474 & 0.250 & 0.328\\
					{}&1:C:H124 (1-1)&4.1&&0.472 & 0.249 & 0.327\\
					{}&Baseline: 2:M:H1 (M-M)&&& 0.257 & 0.919 & 0.402\\
					{}&Baseline: 1:M:H1 (M-M)&&& 0.397 & 0.821 & 0.536\\
					{}&Baseline: 1:C:H1 (1-1)&&& 0.447 & 0.238 & 0.311\\
					{}&Baseline: CP (M-M) &&& 0.230 & 0.926 & 0.368\\
					{}&Baseline: IC (M-M) &&&0.612 & 0.078 & 0.138\\
					\hline
					\multirow{11}{*}{spa-eng-por}
					{}&3:S:H34 (M-M) &3.01&1& 0.368 & 0.870 & 0.517\\
					{}&2:S:H34 (M-M) &3.01&0.49& 0.467 & 0.751 & 0.576\\
					{}&1:S:H14 (M-M) &3.21&0.66& 0.569 & 0.765 & 0.653\\
					{}&3:C:H34 (1-1)&3.01&&0.716 & 0.367 & 0.486\\
					{}&2:C:H34 (1-1)&3.01&&0.716 & 0.367 & 0.486\\
					{}&1:C:H14 (1-1)&3.21&&0.717 & 0.367 & 0.486\\
					{}&Baseline: 2:M:H1 (M-M)&&& 0.389 & 0.870 & 0.537\\
					{}&Baseline: 1:M:H1 (M-M)&&& 0.538 & 0.818 & 0.649\\
					{}&Baseline: 1:C:H1 (1-1)&&& 0.695 & 0.356 & 0.471\\
					{}&Baseline: CP (M-M) &&& 0.368 & 0.870 & 0.517\\
					{}&Baseline: IC (M-M) &&&0.708 & 0.402 & 0.513\\
					\hline
					\multirow{11}{*}{deu-eng-ita}
					{}&3:S:H134 (M-M) &6.14&1& 0.320 & 0.630 & 0.425\\
					{}&2:S:H134 (M-M)&6.14&0.85& 0.477 & 0.534 & 0.504\\
					{}&1:S:H14 (M-M)&6.14&0.85& 0.544 & 0.564 & 0.554\\
					{}&3:C:H134 (1-1)&6.14&&0.621 & 0.310 & 0.413\\
					{}&2:C:H134 (1-1)&6.14&&0.621 & 0.310 & 0.413\\
					{}&1:C:H14 (1-1)&6.14&&0.626 & 0.310 & 0.415\\
					{}&Baseline: 2:M:H1 (M-M)&&& 0.377 & 0.627 & 0.471\\
					{}&Baseline: 1:M:H1 (M-M)&&& 0.542 & 0.565 & 0.553\\
					{}&Baseline: 1:C:H1 (1-1)&&& 0.600 & 0.298 & 0.398\\
					{}&Baseline: CP (M-M) &&& 0.320 & 0.630 & 0.424\\
					{}&Baseline: IC (M-M)&&&0.930 & 0.071 & 0.131\\
					\bottomrule
				\end{tabular}
				\begin{tablenotes}
					\item[] $\langle situatedMethod \rangle ::= \langle cycle \rangle ":" \langle method \rangle ":" \langle heuristic\rangle$
					where \textit{cycle}: symmetry assumption cycle where cycle $\geq$ 1, \textit{method}: \textit{C} as a cognate recognition ($CNF_{cognate}$) or \textit{S} as a cognate \& cognate synonym recognition ($CNF_{cognate} + CNF_{cognateSynonym}$) or \textit{M} as a many-to-many approach ($\Omega_2$ \& $\Omega_3)$ in our previous work \cite{NASUTION16.1238}, \textit{heuristic}: an individual or combined heuristics where H1234 means a combination of heuristic 1 (cognate pair coexistence probability), heuristic 2 (missing contribution rate toward cognate pair coexistence), heuristic 3 (polysemy pivot ambiguity rate), and heuristic 4 (cognate form similarity). CP: Cartesian Product; IC: Inverse Consultation \cite{tanaka-94}; 1-1 : one-to-one translation pair results; M-M : many-to-many translation pair results;			
				\end{tablenotes}
			\end{center}
		\end{threeparttable}
	\end{table*}
	To obtain a good strategy when we want to implement the framework, a balance between precision and recall is crucial. We calculate a harmonic mean of precision and recall using the traditional F1-measure or balanced F1-score by weighting the precision and recall equally. Based on user preference and priority, F0.5-score can be used when precision is considered more important, and F2-score can be used when recall is preferred. The results of all four case studies that targeted the threshold yielding the highest F-score are shown in Table 9. For the case study min-ind-zlm, our best yielding M-M result method (2:S:H14) yields 0.4\% higher F-score than our previous best yielding M-M result method (2:M:H1), 3.4\% higher F-score than CP, and 12.9 times higher F-score than IC, while our best yielding 1-1 result method (3:C:H14) yields 1.3\% higher precision than our previous method (1:C:H1). The high F-score of the CP in the case study min-ind-zlm indicates how very closely-related the input languages are. For the case study deu-eng-nld, our best yielding M-M result method (1:S:H124) yields 0.2\% higher F-score than our previous best yielding M-M result method (1:M:H1), 46\% higher F-score than CP, and 2.9 times higher F-score than IC, while our best yielding 1-1 result method (3:C:H14) yields 5.5\% higher precision than our previous method (1:C:H1). For the case study spa-eng-por, our best yielding M-M result method (1:S:H14) yields 0.6\% higher F-score than our previous best yielding M-M result method (1:M:H1), 26.3\% higher F-score than CP, and 27.3\% higher F-score than IC, while our best yielding 1-1 result method (3:C:H34) yields 3.6\% higher precision than our previous method (1:C:H1). For the case study deu-eng-ita, our best yielding M-M result method (1:S:H14) yields 0.2\% higher F-score than our previous best yielding M-M result method (1:M:H1), 30.7\% higher F-score than CP, and 3.2 times higher F-score than IC, while our best yielding 1-1 result method (3:C:H134) yields 3.6\% higher precision than our previous method (1:C:H1).
	\par
	To enrich the bilingual dictionary result for low-resource languages, cognates and cognate synonyms recognition with higher cycles is the best approach. The exact number of cycles can be customized based on the priority and preference as regards the precision-recall trade-off. The cognates and cognate synonyms recognition with one-cycle is recommended for attaining the highest F-score result, since for almost all case studies in our experiments except min-ind-zlm, it always realized the highest F-score.\par
	\begin{table*}[!h]
		\caption{Comparison of The Proposed Methods and The Previous Method: Case Study min-ind-zlm}
		\footnotesize
		\label{tab:12}
		\begin{threeparttable}
			\begin{center}								
				\begin{tabular}{cc|lll|lllllll}
					\toprule
					\multirow{2}{*}{Comparison}&\multirow{2}{*}{Transgraph}&\multicolumn{3}{c|}{Previous Method}&\multicolumn{6}{c}{Proposed Method}\\
					{}&{}&Precision&Recall&F-score&Precision&Diff.&Recall&Diff.&F-score&Diff.\\
					\hline
					\multirow{15}{*}{1-1\tnote{*}}&0-24&0.92&0.548&0.687&0.92&0&0.548&0&0.687&0\\
					{}&25-40&0.813&0.542&0.65&0.813&0&0.542&0&0.65&0\\
					{}&41-56&0.813&0.52&0.634&0.875&+0.063&0.56&+0.04&0.683&+0.049\\
					{}&57-72&1&0.516&0.681&1&0&0.516&0&0.681&0\\
					{}&73-88&0.9&0.621&0.735&0.9&0&0.621&0&0.735&0\\
					{}&89-104&0.889&0.471&0.615&0.889&0&0.471&0&0.615&0\\
					{}&105-120&0.63&0.447&0.523&0.667&+0.037&0.474&+0.026&0.554&+0.031\\
					{}&121-136&0.552&0.533&0.542&0.552&0&0.533&0&0.542&0\\
					{}&137-152&0.828&0.5&0.623&0.862&+0.034&0.521&+0.021&0.649&+0.026\\
					{}&153-168&0.966&0.346&0.509&1&+0.034&0.358&+0.012&0.527&+0.018\\
					{}&169-184&1&0.352&0.52&1&0&0.352&0&0.52&0\\
					{}&185-200&1&0.34&0.508&1&0&0.34&0&0.508&0\\
					{}&201-216&0.975&0.312&0.473&0.975&0&0.312&0&0.473&0\\
					{}&217-232&0.889&0.294&0.442&0.889&0&0.294&0&0.442&0\\
					{}&233-248&0.866&0.179&0.296&0.878&+0.012&0.181&+0.003&0.301&+0.004\\
					\hline
					\multirow{15}{*}{M-M\tnote{**}}&0-24&0.913&1&0.955&0.913&0&1&0&0.955&0\\
					{}&25-40&0.75&1&0.857&0.75&0&1&0&0.857&0\\
					{}&41-56&0.781&1&0.877&0.781&0&1&0&0.877&0\\
					{}&57-72&0.969&1&0.984&0.969&0&1&0&0.984&0\\
					{}&73-88&0.725&1&0.841&0.725&0&1&0&0.841&0\\
					{}&89-104&0.85&1&0.919&0.864&+0.014&1&0&0.927&+0.008\\
					{}&105-120&0.644&1&0.784&0.644&0&1&0&0.784&0\\
					{}&121-136&0.492&1&0.659&0.492&0&1&0&0.659&0\\
					{}&137-152&0.774&1&0.873&0.774&0&1&0&0.873&0\\
					{}&153-168&0.92&1&0.959&0.92&0&1&0&0.959&0\\
					{}&169-184&0.938&1&0.968&0.938&0&1&0&0.968&0\\
					{}&185-200&0.906&0.99&0.946&0.906&0&0.99&0&0.946&0\\
					{}&201-216&0.886&0.992&0.936&0.886&0&0.992&0&0.936&0\\
					{}&217-232&0.744&0.985&0.848&0.772&+0.028&0.949&-0.037&0.851&+0.003\\
					{}&233-248&0.544&0.864&0.667&0.544&0&0.864&0&0.667&0\\
					\bottomrule
				\end{tabular}	
				\begin{tablenotes}
					\item[*] Comparison between the previous method (1:C:H1 / $\Omega_1$) \cite{NASUTION16.1238} and the proposed method (2:C:H14) which yield one-to-one translation pair results.				
					\item[**] Comparison between the previous method (2:M:H1 / $\Omega_3$) \cite{NASUTION16.1238} and the proposed method (2:S:H14) which yield many-to-many translation pair results.						
				\end{tablenotes}
			\end{center}
		\end{threeparttable}
	\end{table*}
	\begin{table*}[!h]
		\caption{Comparison of The Proposed Methods and The Previous Method: Case Study deu-eng-nld}
		\footnotesize
		\label{tab:13}
		\begin{threeparttable}
			\begin{center}								
				\begin{tabular}{cc|lll|lllllll}
					\toprule
					\multirow{2}{*}{Comparison}&\multirow{2}{*}{Transgraph}&\multicolumn{3}{c|}{Previous Method}&\multicolumn{6}{c}{Proposed Method}\\
					{}&{}&Precision&Recall&F-score&Precision&Diff.&Recall&Diff.&F-score&Diff.\\
					\hline
					\multirow{15}{*}{1-1\tnote{*}}&0-16&0.529&0.9&0.667&0.588&+0.059&1&+0.1&0.741&+0.074\\
					{}&17-34&0.478&0.478&0.478&0.522&+0.043&0.522&+0.043&0.522&+0.043\\
					{}&35-52&0.594&0.463&0.521&0.594&0&0.463&0&0.521&0\\
					{}&53-70&0.286&0.25&0.267&0.286&0&0.25&0&0.267&0\\
					{}&71-88&0.406&0.271&0.325&0.5&+0.094&0.333&+0.063&0.4&+0.075\\
					{}&89-106&0.447&0.333&0.382&0.447&0&0.333&0&0.382&0\\
					{}&107-124&0.641&0.321&0.427&0.667&+0.026&0.333&+0.013&0.444&+0.017\\
					{}&125-142&0.455&0.235&0.31&0.455&0&0.235&0&0.31&0\\
					{}&143-160&0.439&0.22&0.293&0.512&+0.073&0.256&+0.037&0.341&+0.049\\
					{}&161-178&0.333&0.237&0.277&0.426&+0.093&0.303&+0.066&0.354&+0.077\\
					{}&179-196&0.526&0.265&0.353&0.517&-0.009&0.265&0&0.351&-0.002\\
					{}&197-214&0.435&0.195&0.269&0.435&0&0.195&0&0.269&0\\
					{}&215-232&0.408&0.228&0.293&0.38&-0.028&0.213&-0.016&0.273&-0.02\\
					{}&233-250&0.41&0.211&0.279&0.457&+0.047&0.23&+0.019&0.306&+0.027\\
					{}&251-268&0.446&0.135&0.208&0.485&+0.039&0.14&+0.004&0.217&+0.009\\
					\hline
					\multirow{15}{*}{M-M\tnote{**}}&0-16&0.417&1&0.588&0.417&0&1&0&0.588&0\\
					{}&17-34&0.435&0.87&0.58&0.455&+0.02&0.87&0&0.597&+0.017\\
					{}&35-52&0.559&0.927&0.697&0.587&+0.028&0.902&-0.024&0.712&+0.014\\
					{}&53-70&0.329&0.844&0.474&0.329&0&0.844&0&0.474&0\\
					{}&71-88&0.392&0.833&0.533&0.392&0&0.833&0&0.533&0\\
					{}&89-106&0.349&0.882&0.5&0.366&+0.017&0.804&-0.078&0.503&+0.003\\
					{}&107-124&0.531&0.987&0.691&0.531&0&0.987&0&0.691&0\\
					{}&125-142&0.363&0.729&0.484&0.389&+0.026&0.659&-0.071&0.489&+0.005\\
					{}&143-160&0.371&0.915&0.528&0.371&0&0.915&0&0.528&0\\
					{}&161-178&0.274&0.961&0.427&0.304&+0.029&0.763&-0.197&0.434&+0.008\\
					{}&179-196&0.33&0.947&0.49&0.33&0&0.947&0&0.49&0\\
					{}&197-214&0.254&0.675&0.369&0.287&+0.033&0.643&-0.032&0.397&+0.027\\
					{}&215-232&0.224&0.898&0.358&0.271&+0.047&0.646&-0.252&0.381&+0.023\\
					{}&233-250&0.197&0.87&0.322&0.254&+0.056&0.671&-0.199&0.368&+0.046\\
					{}&251-268&0.199&0.849&0.323&0.301&+0.102&0.561&-0.288&0.392&+0.069\\
					\bottomrule
				\end{tabular}	
				\begin{tablenotes}
					\item[*] Comparison between the previous method (1:C:H1 / $\Omega_1$) \cite{NASUTION16.1238} and the proposed method (2:C:H14) which yield one-to-one translation pair results.				
					\item[**] Comparison between the previous method (2:M:H1 / $\Omega_3$) \cite{NASUTION16.1238} and the proposed method (2:S:H14) which yield many-to-many translation pair results.							
				\end{tablenotes}
			\end{center}
		\end{threeparttable}
	\end{table*}
	\begin{table*}[!h]
		\caption{Comparison of The Proposed Methods and The Previous Method: Case Study spa-eng-por}
		\footnotesize
		\label{tab:14}
		\begin{threeparttable}
			\begin{center}								
				\begin{tabular}{cc|lll|lllllll}
					\toprule
					\multirow{2}{*}{Comparison}&\multirow{2}{*}{Transgraph}&\multicolumn{3}{c|}{Previous Method}&\multicolumn{6}{c}{Proposed Method}\\
					{}&{}&Precision&Recall&F-score&Precision&Diff.&Recall&Diff.&F-score&Diff.\\
					\hline
					\multirow{15}{*}{1-1\tnote{*}}&0-24&1&0.833&0.909&1&0&0.833&0&0.909&0\\
					{}&25-45&0.714&0.75&0.732&0.714&0&0.75&0&0.732&0\\
					{}&46-66&0.81&0.68&0.739&0.81&0&0.68&0&0.739&0\\
					{}&67-87&0.762&0.421&0.542&0.762&0&0.421&0&0.542&0\\
					{}&88-108&0.667&0.467&0.549&0.714&+0.048&0.5&+0.033&0.588&+0.039\\
					{}&109-129&0.762&0.471&0.582&0.81&+0.048&0.5&+0.029&0.618&+0.036\\
					{}&130-150&0.64&0.364&0.464&0.68&+0.04&0.386&+0.023&0.493&+0.029\\
					{}&151-171&0.724&0.382&0.5&0.69&-0.034&0.364&-0.018&0.476&-0.024\\
					{}&172-192&0.9&0.351&0.505&0.9&0&0.351&0&0.505&0\\
					{}&193-213&0.6&0.296&0.397&0.6&0&0.296&0&0.397&0\\
					{}&214-234&0.61&0.212&0.314&0.585&-0.024&0.203&-0.008&0.302&-0.013\\
					{}&235-255&0.587&0.287&0.386&0.63&+0.043&0.309&+0.021&0.414&+0.029\\
					{}&256-276&0.577&0.288&0.385&0.615&+0.038&0.308&+0.019&0.41&+0.026\\
					{}&277-297&0.678&0.276&0.392&0.712&+0.034&0.29&+0.014&0.412&+0.02\\
					{}&298-318&0.708&0.221&0.337&0.74&+0.031&0.231&+0.01&0.352&+0.015\\
					\hline
					\multirow{15}{*}{M-M\tnote{**}}&0-24&1&0.833&0.909&1&0&0.833&0&0.909&0\\
					{}&25-45&0.714&0.75&0.732&0.714&0&0.75&0&0.732&0\\
					{}&46-66&0.75&0.84&0.792&0.75&0&0.84&0&0.792&0\\
					{}&67-87&0.667&0.737&0.7&0.667&0&0.737&0&0.7&0\\
					{}&88-108&0.585&0.8&0.676&0.585&0&0.8&0&0.676&0\\
					{}&109-129&0.596&0.824&0.691&0.596&0&0.824&0&0.691&0\\
					{}&130-150&0.667&0.909&0.769&0.678&+0.011&0.909&0&0.777&+0.007\\
					{}&151-171&0.632&0.782&0.699&0.646&+0.014&0.764&-0.018&0.7&+0.001\\
					{}&172-192&0.663&0.766&0.711&0.663&0&0.766&0&0.711&0\\
					{}&193-213&0.46&0.704&0.556&0.46&0&0.704&0&0.556&0\\
					{}&214-234&0.438&0.534&0.481&0.458&+0.021&0.508&-0.025&0.482&+0.001\\
					{}&235-255&0.433&0.83&0.569&0.433&0&0.83&0&0.569&0\\
					{}&256-276&0.359&0.817&0.499&0.359&0&0.817&0&0.499&0\\
					{}&277-297&0.36&0.862&0.508&0.433&+0.073&0.697&-0.166&0.534&+0.026\\
					{}&298-318&0.255&0.779&0.384&0.359&+0.104&0.59&-0.189&0.446&+0.062\\
					\bottomrule
				\end{tabular}			
				\begin{tablenotes}
					\item[*] Comparison between the previous method (1:C:H1 / $\Omega_1$) \cite{NASUTION16.1238} and the proposed method (2:C:H14) which yield one-to-one translation pair results.				
					\item[**] Comparison between the previous method (2:M:H1 / $\Omega_3$) \cite{NASUTION16.1238} and the proposed method (2:S:H14) which yield many-to-many translation pair results.								
				\end{tablenotes}
			\end{center}
		\end{threeparttable}
	\end{table*}
	\begin{table*}[!h]
		\caption{Comparison of The Proposed Methods and The Previous Method: Case Study deu-eng-ita}
		\footnotesize
		\label{tab:15}
		\begin{threeparttable}
			\begin{center}								
				\begin{tabular}{cc|lll|lllllll}
					\toprule
					\multirow{2}{*}{Comparison}&\multirow{2}{*}{Transgraph}&\multicolumn{3}{c|}{Previous Method}&\multicolumn{6}{c}{Proposed Method}\\
					{}&{}&Precision&Recall&F-score&Precision&Diff.&Recall&Diff.&F-score&Diff.\\
					\hline
					\multirow{15}{*}{1-1\tnote{*}}&0-34&0.943&0.367&0.528&0.943&0&0.367&0&0.528&0\\
					{}&35-64&0.633&0.235&0.342&0.7&+0.067&0.259&+0.025&0.378&+0.036\\
					{}&65-94&0.7&0.3&0.42&0.667&-0.033&0.286&-0.014&0.4&-0.02\\
					{}&95-124&0.533&0.246&0.337&0.667&+0.133&0.308&+0.062&0.421&+0.084\\
					{}&125-154&0.667&0.274&0.388&0.667&0&0.274&0&0.388&0\\
					{}&155-184&0.5&0.167&0.25&0.533&+0.033&0.178&+0.011&0.267&+0.017\\
					{}&185-214&0.567&0.23&0.327&0.633&+0.067&0.257&+0.027&0.365&+0.038\\
					{}&215-244&0.646&0.316&0.425&0.667&+0.021&0.327&+0.01&0.438&+0.014\\
					{}&245-274&0.694&0.256&0.374&0.673&-0.02&0.248&-0.008&0.363&-0.011\\
					{}&275-304&0.689&0.341&0.456&0.711&+0.022&0.352&+0.011&0.471&+0.015\\
					{}&305-334&0.556&0.197&0.291&0.587&+0.031&0.213&+0.016&0.312&+0.021\\
					{}&335-364&0.561&0.182&0.275&0.542&-0.019&0.182&0&0.272&-0.002\\
					{}&365-394&0.54&0.177&0.267&0.556&+0.016&0.182&+0.005&0.275&+0.008\\
					{}&395-424&0.519&0.169&0.256&0.532&+0.013&0.174&+0.004&0.262&+0.006\\
					{}&425-454&0.544&0.184&0.275&0.562&+0.017&0.189&+0.005&0.283&+0.007\\
					\hline
					\multirow{15}{*}{M-M\tnote{**}}&0-34&0.946&0.389&0.551&0.946&0&0.389&0&0.551&0\\
					{}&35-64&0.672&0.481&0.561&0.672&0&0.481&0&0.561&0\\
					{}&65-94&0.627&0.529&0.574&0.627&0&0.529&0&0.574&0\\
					{}&95-124&0.593&0.538&0.565&0.593&0&0.538&0&0.565&0\\
					{}&125-154&0.61&0.493&0.545&0.61&0&0.493&0&0.545&0\\
					{}&155-184&0.583&0.389&0.467&0.583&0&0.389&0&0.467&0\\
					{}&185-214&0.633&0.514&0.567&0.633&0&0.514&0&0.567&0\\
					{}&215-244&0.515&0.52&0.518&0.515&0&0.52&0&0.518&0\\
					{}&245-274&0.535&0.406&0.462&0.535&0&0.406&0&0.462&0\\
					{}&275-304&0.455&0.549&0.498&0.455&0&0.549&0&0.498&0\\
					{}&305-334&0.444&0.441&0.443&0.444&0&0.441&0&0.443&0\\
					{}&335-364&0.407&0.409&0.408&0.419&+0.012&0.398&-0.011&0.408&0\\
					{}&365-394&0.367&0.438&0.399&0.401&+0.034&0.422&-0.016&0.411&+0.012\\
					{}&395-424&0.331&0.462&0.386&0.379&+0.048&0.419&-0.042&0.398&+0.013\\
					{}&425-454&0.226&0.488&0.309&0.339&+0.112&0.336&-0.152&0.338&+0.028\\
					\bottomrule
				\end{tabular}			
				\begin{tablenotes}
					\item[*] Comparison between the previous method (1:C:H1 / $\Omega_1$) \cite{NASUTION16.1238} and the proposed method (2:C:H14) which yield one-to-one translation pair results.				
					\item[**] Comparison between the previous method (2:M:H1 / $\Omega_3$) \cite{NASUTION16.1238} and the proposed method (2:S:H14) which yield many-to-many translation pair results.					
				\end{tablenotes}
			\end{center}
		\end{threeparttable}
	\end{table*}
	For the case study deu-eng-nld, the best one-to-one cognate (3:C:H14) method precision is unexpectedly low, 0.474 while the lower language similarity case studies (spa-eng-por and deu-eng-ita) with the same cycle have higher precision (0.716 and 0.621 respectively). The case study deu-eng-nld always yielded lower F-scores than case studies deu-eng-ita and spa-eng-por when the methods that generate many-to-many results were applied. We believe that inadequacy of the gold standard was the cause of this counter-intuitive result. For the case study deu-eng-nld, if we look at the ratio of the size of the Cartesian product across transgraph in Table 6 and the size of the combined dictionaries for evaluation in Table 5, relative to the ratio of the gold standard and the Cartesian product within the transgraph, it is obvious that the ratio is inadequate compared to the other case study languages.
	
	\subsubsection{Statistical Significant Test}
	To show that our proposed methods are statistically significant compared to our previous methods \cite{NASUTION16.1238}, as listed in Table 10-13, for each case study, firstly, we split the dataset into several datapoints (transgraphs), then we compare the potentially best methods yielding the most many-to-many translation pairs (M-M), i.e., the 2:S:H14 to our previous method that potentially yielding the most many-to-many translation pairs (M-M), i.e., 2:M:H1. We also compare the potentially best methods yielding the most one-to-one translation pairs (1-1), i.e., the 2:C:H14 to our previous method that yielding one-to-one translation pairs (1-1), i.e., 1:C:H1. Student's paired t-test is a good statistical procedure used in Information Retrieval research to determine whether the mean difference between two sets of observations is zero \cite{Smucker2007}. It is very useful to show that our proposed methods are truly better than our previous methods rather than performed better by chance. In a student's paired t-test, each subject or entity is measured twice, resulting in pairs of observations. In this paper, we use the same set of datapoints and conduct the student's paired t-test with precision and F-score as measures. Since we expect that our proposed methods have improvement compared to our previous methods, we choose a one-tailed t-test. There are two sets of null hypotheses (precision null hypotheses and F-score null hypotheses), which are that the true precision or F-score means difference between the proposed methods and our previous methods are equal to zero. We decide 0.05 cutoff value for determining statistical significance which corresponds to a 5\% (or less) chance of obtaining a result like the one that was observed if the null hypotheses were true. For all case studies min-ind-zlm, deu-eng-nld, spa-eng-por, and deu-eng-ita, we reject the precision null hypotheses since the p-value of the tests are 0.00732, 0.00007, 0.00398, 0.00464, respectively, which are all smaller than 0.05. For all case studies min-ind-zlm, deu-eng-nld, spa-eng-por, and deu-eng-ita, we also reject the F-score null hypotheses since the p-value of the tests are 0.01673, 0.00034, 0.00652, 0.00783, respectively, which are all smaller than 0.05. Thus, our proposed methods have statistically significant improvement of precision and F-score compared to our previous methods.
	
	\subsubsection{Hyperparameter Optimization}
	We have shown that our methods outperformed the baselines in the previous sections. Nevertheless, before implementing our model in a big scale, we need to validate how good our model perform in practice with unknown data. Since there is not enough data available to partition it into separate training and test sets without losing significant modelling or testing capability, a good way to properly estimate model prediction performance is to use cross-validation as a powerful general technique. Due to the computational complexity of our model, we conduct 3-folds cross validation to predict the optimal hyperparameters (cognate threshold and cognate synonym threshold) to gain the highest F-score as shown in Table 14. We optimize the hyperparameters with a grid search by incrementing the cognate threshold from 0 to the highest cost of violating the constraints with 0.01 intervals and incrementing the cognate synonym threshold from 0 to 1 with 0.01 intervals in order to find the highest F-score. We choose the same methods as in Table 10-13, the potentially best methods yielding the most one-to-one translation pairs (1-1), i.e., the 2:C:H14 and the potentially best methods yielding the most many-to-many translation pairs (M-M), i.e., the 2:S:H14. For all case studies, the mean F-score approaches the mean F-score of the overfitting model in Table 10-13.
	\begin{table*}[!h]
		\caption{Cognate Threshold and Cognate Synonym Threshold Optimization}
		\scriptsize
		\label{tab:14}
		\begin{center}								
			\begin{tabular}{ccl|ll|lllllll}
				\toprule
				\multirow{2}{*}{Case Study}&\multirow{2}{*}{Method}&\multirow{2}{*}{Validation Set}&\multicolumn{2}{c|}{Optimal Threshold}&\multicolumn{5}{c}{Testing on Unknown Data}\\
				{}&{}&{}&Cognate&Cognate Synonym&Test Set&Precision&Recall&F-score&Mean F-score\\
				\hline
				\multirow{6}{*}{min-ind-zlm}&\multirow{3}{*}{2CH14}&0-82, 83-165&1.35&-&166-248&0.933&0.257&0.403&\multirow{3}{*}{0.559}\\
				{}&{}&0-82, 166-248&4.79&-&83-165&0.786&0.471&0.589&{}\\
				{}&{}&83-165, 166-248&4.79&-&0-82&0.916&0.547&0.685&{}\\
				{}&\multirow{3}{*}{2SH14}&0-82, 83-165&1.99&1&166-248&0.688&0.933&0.792&\multirow{3}{*}{0.853}\\
				{}&{}&0-82, 166-248&4.79&0.26&83-165&0.729&1&0.843&{}\\
				{}&{}&83-165, 166-248&4.79&0.26&0-82&0.858&1&0.924&{}\\
				\hline
				\multirow{6}{*}{deu-eng-nld}&\multirow{3}{*}{2CH14}&0-90, 91-179&1.85&-&180-268&0.467&0.185&0.265&\multirow{3}{*}{0.359}\\
				{}&{}&0-90, 180-268&1.97&-&91-179&0.493&0.285&0.361&{}\\
				{}&{}&91-179, 180-268&1.97&-&0-90&0.485&0.423&0.452&{}\\
				{}&\multirow{3}{*}{2SH14}&0-90, 91-179&1.85&1&180-268&0.219&0.86&0.35&\multirow{3}{*}{0.474}\\
				{}&{}&0-90, 180-268&1.97&0.51&91-179&0.361&0.893&0.514&{}\\
				{}&{}&91-179, 180-268&1.97&0.51&0-90&0.41&0.878&0.559&{}\\
				\hline
				\multirow{6}{*}{spa-eng-por}&\multirow{3}{*}{2CH14}&0-106, 107-212&2.96&-&213-318&0.676&0.268&0.384&\multirow{3}{*}{0.525}\\
				{}&{}&0-106, 213-318&3.21&-&107-212&0.724&0.366&0.486&{}\\
				{}&{}&107-212, 213-318&3.21&-&0-106&0.804&0.628&0.705&{}\\
				{}&\multirow{3}{*}{2SH14}&0-106, 107-212&2.96&0.51&213-318&0.394&0.636&0.487&\multirow{3}{*}{0.639}\\
				{}&{}&0-106, 213-318&3.21&0.51&107-212&0.603&0.756&0.671&{}\\
				{}&{}&107-212, 213-318&3.21&0.51&0-106&0.719&0.803&0.759&{}\\
				\hline
				\multirow{6}{*}{deu-eng-ita}&\multirow{3}{*}{2CH14}&0-150, 151-302&1.5&-&303-454&0.557&0.202&0.297&\multirow{3}{*}{0.371}\\
				{}&{}&0-150, 303-454&6.14&-&151-302&0.652&0.279&0.391&{}\\
				{}&{}&151-302, 303-454&6.14&-&0-150&0.735&0.3&0.426&{}\\
				{}&\multirow{3}{*}{2SH14}&0-150, 151-302&1.5&0.01&303-454&0.341&0.414&0.374&\multirow{3}{*}{0.479}\\
				{}&{}&0-150, 303-454&6.14&0.56&151-302&0.531&0.481&0.505&{}\\
				{}&{}&151-302, 303-454&6.14&0.56&0-150&0.67&0.478&0.558&{}\\
				\bottomrule
			\end{tabular}							
		\end{center}
	\end{table*}

\clearpage

	\section{Conclusion}
	Our strategy to create high quality many-to-many translation pairs between closely-related languages consists of two steps. We first recognize cognates from direct and indirect connectivity via pivot word(s) by iterating multiple symmetry assumption cycles to reach more cognates in the transgraph. Once we obtain a list of cognates, the next step identifies synonyms of those cognates.\par
	The result of case studies showed that our method offers good performance on weakly related high-resource languages. Thus, our method has the potential to complement other bilingual dictionary creation methods like word alignment models using parallel corpora. Our method shows particularly high performance on the closely related low-resource language case study. Our proposed methods have statistically significant improvement of precision and F-score compared to our previous methods in spite of sacrificing the recall a little bit.\par
	Our key research contribution is a generalized constraint-based bilingual lexicon induction framework for closely related low-resource languages. This generalization makes our method applicable for a wider range of language groups than the one-to-one approach. Our customizable approach allows the user to conduct cross validation to predict the optimal hyperparameters (cognate threshold and cognate synonym threshold) with various combination of heuristics and number of symmetry assumption cycles to gain the highest F-score. To the best of our knowledge, our study is the first attempt to recognize both cognates and cognate synonyms in bilingual lexicon induction. \par

	\begin{acks}
		This research was partially supported by a Grant-in-Aid for Scientific Research (A) (17H00759, 2017-2020) and a Grant-in-Aid for Young Scientists (A) (17H04706, 2017-2020) from Japan Society for the Promotion of Science (JSPS). The first author was supported by Indonesia Endownment Fund for Education (LPDP).
	\end{acks}

	\bibliographystyle{ACM-Reference-Format-Journals}
	\bibliography{tallip}
	
	\renewcommand{\shortauthors}{A. H. Nasution et al.}	
\end{document}